\documentclass[11pt]{article}

\usepackage[preprint]{acl}

\usepackage{times}
\usepackage{latexsym}

\usepackage[T1]{fontenc}

\usepackage[utf8]{inputenc}

\usepackage{microtype}

\usepackage{inconsolata}

\usepackage{graphicx}

\usepackage{amsmath}
\usepackage{amsfonts}   
\usepackage{booktabs}
\usepackage{tabularx}
\usepackage{graphicx} 
\usepackage{makecell}  
\usepackage{enumitem}
\usepackage{booktabs}
\usepackage{ragged2e}
\usepackage{multirow}
\usepackage{textcomp}
\usepackage{float}
\usepackage[table]{xcolor}
\usepackage{amssymb}
\usepackage{pifont}

\usepackage{hyperref}
\usepackage{url}
\usepackage{xurl}

\usepackage{xspace}
\usepackage{graphicx}
\usepackage{amsmath}
\usepackage{amssymb}
\usepackage{mathtools}
\usepackage{amsthm}
\usepackage{bbm}
\usepackage{multirow}
\usepackage{booktabs}
\usepackage[most]{tcolorbox}
\usepackage{fontawesome}
\usepackage{algorithm}
\usepackage{algpseudocode}
\usepackage{amsmath,amssymb}
\usepackage{microtype}
\algnewcommand\Input{\item[\textbf{Input:}]}
\algnewcommand\Output{\item[\textbf{Output:}]}

\newcommand\blfootnote[1]{%
\begingroup
\renewcommand\thefootnote{}\footnote{#1}%
\addtocounter{footnote}{-1}%
\endgroup
}

\newcommand{\methodname}{\textsc{AttnPO}\xspace}
\title{\methodname: Attention-Guided Process Supervision for Efficient Reasoning}

\author{
    Shuaiyi Nie$^{1,2}$,
    Siyu Ding$^{3^\ddagger}$,
    Wenyuan Zhang$^{1,2}$,
    Linhao Yu$^4$,
    Tianmeng Yang$^{3}$,
    Yao Chen$^{1,2}$,\\
    \textbf{Weichong Yin$^{3}$,}
    \textbf{Yu Sun$^{3}$,}
    \textbf{Hua Wu$^{3}$,}
    \textbf{Tingwen Liu$^{1,2^\dagger}$}\\
    $^1$Institute of Information Engineering, Chinese Academy of Sciences \\
    $^2$School of Cyber Security, University of Chinese Academy of Sciences \\
    $^3$Baidu Inc.
    $^4$Tianjin University\\
   \texttt{\{nieshuaiyi, liutingwen\}@iie.ac.cn}\\[4pt]
}

\begin{document}
\maketitle

\begin{abstract}

Large reasoning models trained with reinforcement learning and verifiable rewards (RLVR) achieve strong performance on complex reasoning tasks, yet often overthink, generating redundant reasoning without performance gains.
Existing trajectory-level length penalties often fail to effectively shorten  reasoning length and degrade accuracy, as they uniformly treat all reasoning steps and lack fine-grained signals to distinguish redundancy from necessity.
Meanwhile, process-supervised methods are typically resource-intensive and suffer from inaccurate credit assignment.
To address these issues, we propose \methodname, a low-overhead process-supervised RL framework that leverages the model’s intrinsic attention signals for step-level credit assignment.
We first identify a set of special attention heads that naturally focus on essential steps while suppressing redundant ones.
By leveraging the attention scores of these heads, We then employ two sub-strategies to mitigate overthinking by discouraging redundant steps while preserving accuracy by reducing penalties on essential steps.
Experimental results show that \methodname substantially reduces reasoning length while significantly improving performance across 9 benchmarks\footnote{Code link: \url{https://github.com/NieSYsc20/AttnPO}.}.
\end{abstract}

\blfootnote{
$^\dagger$Corresponding author. 
$^\ddagger$ Project lead.
}

\vspace{-2em}
\section{Introduction}

Recent advances in Large Reasoning Models (LRMs)~\cite{guo2025deepseek,openai-o1} have demonstrated remarkable performance on complex reasoning tasks. 
This success is largely driven by outcome-supervised RL methods, represented by GRPO~\cite{shao2024deepseekmath}, which omits the critic model in PPO~\cite{schulman2017proximal} by utilizing verifiable outcome rewards to estimate group advantages.
A hallmark of LRMs is their ability to produce deliberative long CoT that employ advanced strategies such as self-reflection and multi-path exploration~\cite{li2025system}.
However, long CoT fosters overthinking~\cite{zhang2025s1bench}: LRMs indiscriminately apply verbose reasoning, wasting computation on even trivial operations.

\begin{figure}[t]
  \centering
  \includegraphics[width=7.5cm]{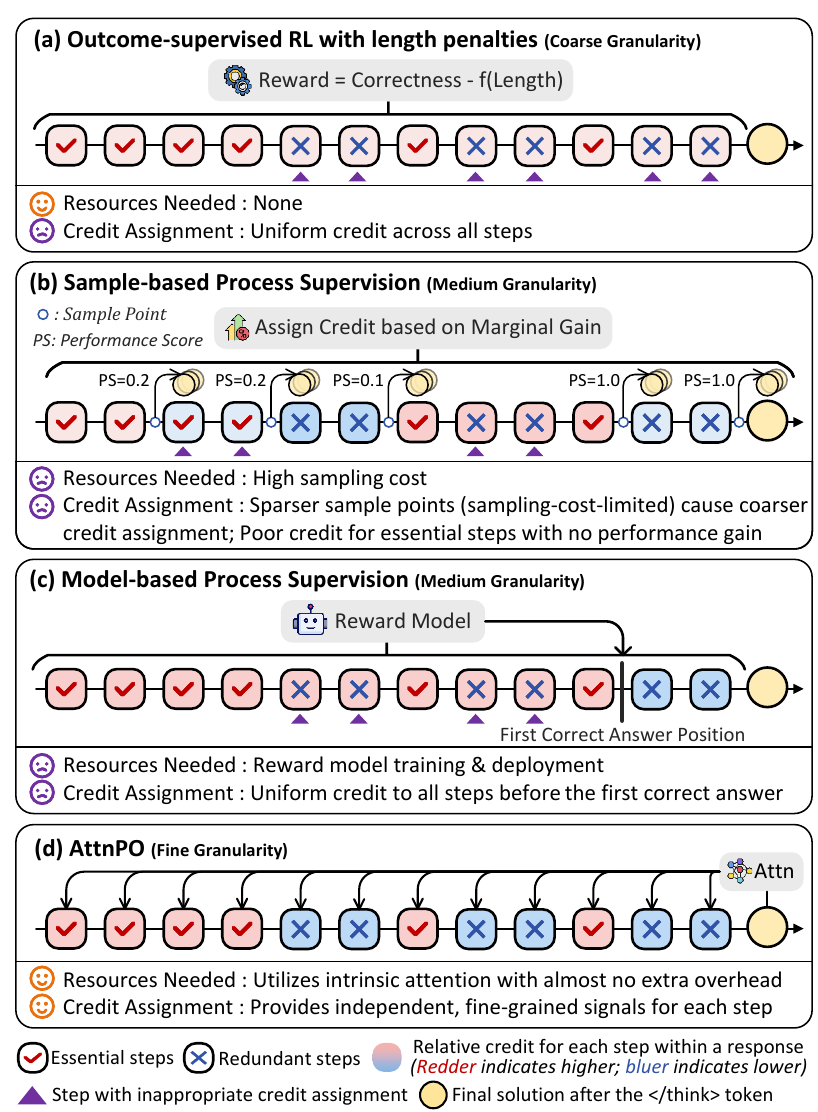}
  \caption{\methodname vs. other reinforcement learning methods for efficient reasoning.}
  \label{fig:motivation}
  \vspace{-1.0em}
\end{figure}

Integrating length penalties in outcome-supervise RL~(Figture~\ref{fig:motivation}(a)) is widely adopted to mitigate overthinking. 
The core idea is to assign higher rewards to responses that are both correct and concise~\cite{arora2025training,team2025kimi,yi2025shorterbetter}. 
However, this coarse-grained feedback assigns uniform credit to all reasoning steps, failing to provide discriminative learning signals to distinguish between necessary and redundant steps.

Therefore, recent studies focus on process supervision to down-weight the credit assigned to redundant steps.
As shown in Figure~\ref{fig:motivation}, sample-based methods~(Figure~\ref{fig:motivation}(b)) estimate the marginal performance gain of intermediate steps via Monte Carlo sampling of immediate final solutions~\cite{dai2025s,yue2025promoting}, while model-based methods~(Figure~\ref{fig:motivation}(c)) train a reward model to pinpoint the first correct answer position, treating only the subsequent steps as redundant~\cite{jiang2025overthinking,tan2025towards}.
However, these methods face two key limitations:
(1) High Overhead, which requires additional sampling or a reward model;
(2) Inaccurate Credit Assignment, which fails to provide independent step-wise signals and may lead to erroneous credit allocation.
A key question thus emerges:
\textbf{Can we, with almost no additional resource costs, rely solely on the model’s intrinsic signals to achieve fine-grained, step-wise supervision of which steps are essential or redundant?}

In this study, we delve into the model’s intrinsic attention mechanism and identify a striking phenomenon: 
during the final answer generation (following the \texttt{</think>} token), a subset of heads naturally focus on essential steps by assigning them higher attention weights while suppressing redundant ones. 
We term these \textbf{Key-Focus Heads} (KFHs).
Building on this finding, our experiments further reveal:
(1) Positional Distribution, KFHs are mainly located in middle-to-late layers;
(2) Training Stability, where the behavior of KFHs varies little under RL training with length penalties;
and (3) Sparse Sufficiency, as a small set of KFHs suffice for distinction, with performance saturating rapidly as more heads are added.

Based on these insights, we propose \textbf{\methodname} (Attention-guided Policy Optimization), a novel RL framework that enables step-level supervision by leveraging KFHs’ attention scores, achieving superior performance with negligible training overhead.
Specifically, \methodname scales the outcome-level advantage in a stepwise manner via two strategies on correct responses:
(1) for correct responses with positive outcome advantages, \methodname attenuates the positive advantage assigned to redundant steps, preventing over-encouragement them and mitigating overthinking; and
(2) for correct responses with negative outcome advantages, \methodname attenuates the negative advantage assigned to essential steps, avoiding excessive penalization of valid reasoning and alleviating performance degradation.

Our contributions are threefold:
(1) We are the first to reveal the existence of KFHs in LRMs, which naturally focus on essential reasoning steps while suppressing redundant ones;
(2) Based on this insight, we propose \methodname, a low-overhead process-supervised RL framework that leverages intrinsic attention signals for step-level credit assignment;
(3) Empirical results demonstrate that \methodname substantially shortens reasoning length while improving accuracy. For example, on DeepSeek-R1-Distill-Qwen-1.5B, \methodname achieves an average +7.3-point absolute accuracy gain with a 60\% reduction in reasoning length across six math benchmarks.

\section{Preliminary}
\label{sec:Preliminary}

\paragraph{Two-stage Generation of LRMs}
Given an input question $q$, an LRM $\mathcal{M}$ generates a response $o = (\mathcal{T}, \mathcal{F})$ in two sequential stages, explicitly separated by a special \texttt{<think>} token.
First, the model produces a deliberate thinking process $\mathcal{T}$, which may contain redundant reasoning structures, such as unnecessary reflection or verification~\cite{jiang2026foe}.
Then, it outputs a concise final solution $\mathcal{F}$ by summarizing $\mathcal{T}$ and retaining only the essential steps required to solve the problem.

\paragraph{Step Segment} 
The thinking process $\mathcal{T}$ can be decomposed into meaningful intermediate steps, whose boundaries are typically marked by specific special phrases that capture key points in reasoning, such as pauses, reflections or shifts in direction~\cite{hammoud2025beyond,choi2025think,yue2025promoting}.
Thus we segment the thinking process with special phrases, which fall into three categories:
Confused phrases (e.g., ``Wait'') indicate the model is entering a reflection, verification, or branching stage;
Progression phrases(e.g., ``First'') signal the continuation of the current reasoning;  
Summary phrases(e.g., ``Therefore'') summarize the preceding reasoning.  
To avoid overly short segments, we adaptively merge segments shorter than a threshold.
During segmentation and merging, we separate segments that begin with confusion phrases from the preceding context, since they often trigger redundant verification or reflection. Isolating them makes such redundancy easier to detect and penalize.
See Appendix~\ref{sec:Step_Segmentation_Algorithm} for details.

\begin{figure*}[!htbp] 
\centering  
\includegraphics[width=16cm]{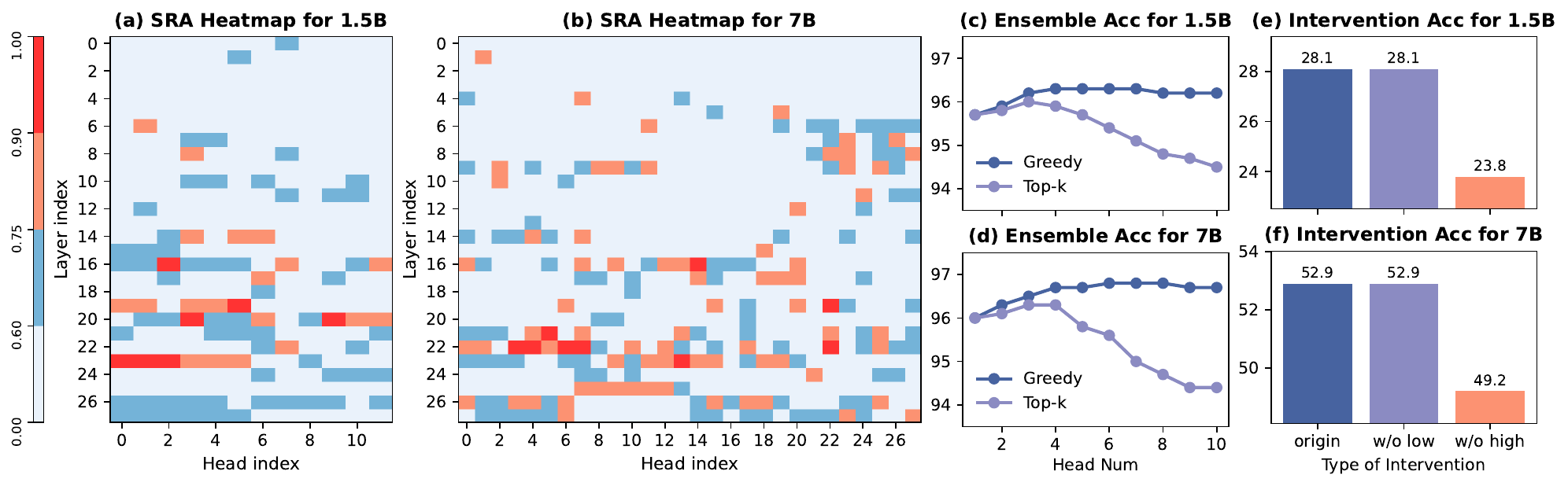}
\caption{Probing results of Key-Focus Heads.}
\label{fig:KFHs}
\vspace{-1em}
\end{figure*}

\section{Key-Focus Heads}
Following the two-stage generation paradigm described in~\ref{sec:Preliminary}, LRMs transform a deliberative thinking process—often containing redundant reasoning—into a concise final solution, which requires non-trivial selection and aggregation of essential information.
%
%
In Transformer architecture, attention serves as the primary mechanism for information selection~\cite{vaswani2017attention}, 
and prior work has shown that different attention heads specialize in distinct functions~\cite{zheng2024attention, li2025cai, chen2026sparse}.
Based on this observation, we hypothesize that \textbf{during final-answer generation, there exist a specific subset of attention heads that selectively focus on essential reasoning steps while ignoring redundant ones, thereby summarizing and integrating the thinking process.} 
To test this hypothesis, we examine the attention distributions over essential and redundant steps across different attention heads.

\subsection{Probing Setup}

\paragraph{Probe Data Construction} 
To analyze how different attention heads attend to essential versus redundant steps, we require a probing dataset with reliable step-level annotations.
We therefore sample 300 questions from Deepscaler~\cite{deepscaler2025} that are not difficult for DeepSeek-R1-Distill-Qwen-1.5B~(average accuracy $\geq 0.875$ over eight samples).
This restriction is important because, on harder problems, LRMs often exhibit exploratory reasoning. Such reasoning may appear redundant, but still contributes to correct solutions, making reliable step annotation difficult and thereby affecting the head identification.
%
%
For each question, we sample eight responses, select the median-length correct one, and segment its thinking process following Section~\ref{sec:Step_Segmentation_Algorithm}.
We then annotate each reasoning step using three state-of-the-art LLMs\footnote{We use GPT-5, Claude-4-sonnet, and Gemini-2.5-pro.}, and retain only the steps on which all three models agree on the label, thereby reducing annotation noise. 
Thus, for each thinking process $\mathcal{T}$, we obtain a set of essential steps $E_s$ and a set of redundant steps $R_s$.
Details are provided in~\ref{sec:Probe_Data_Construction}.

\paragraph{Attention-Based Step Scores} 
For each head at layer $l$ and head $h$, we compute an attention-based step score
$\mathcal{S}^{l,h}_{s_k}$ for each reasoning step $s_k$, where $k$ indexes the step index,
based on the attention weights used during final-answer generation.
Specifically, let $a_{m\to n}^{l,h}$ denote the attention weight from token $m$ in the final solution $\mathcal{F}$ to token $n$ in step $s_k$.
We define the step score $\mathcal{S}^{l,h}_{s_k}$ as:
\begin{equation}
\label{stepscore}
\mathcal{S}^{l,h}_{s_k} = \frac{1}{|s_k|} \sum_{m \in \mathcal{F}} \sum_{n \in s_k} a_{m\to n}^{l,h}.
\end{equation}
This score measures the average attention received by each token in step $s_k$ from the final solution.

\paragraph{Evaluation Metrics: Step Ranking Accuracy}
We evaluate how well each attention head distinguishes between essential and redundant steps.
For each thinking process $\mathcal{T}$, we form all essential–redundant step pairs
$\mathcal{P}_{\mathcal{T}} = \{(s^{+}, s^{-}) \mid s^{+} \in E_{s},\, s^{-} \in R_{s}\}$.
For head $h$ at layer $l$, we define Step Ranking Accuracy (SRA) as:
\begin{equation}
\mathrm{SRA}_{l,h}
= \frac{|\{(s^{+}, s^{-}) \in \mathcal{P}_{\mathcal{T}} \mid \mathcal{S}^{l,h}_{s^+} > \mathcal{S}^{l,h}_{s^-}\}|}{|\mathcal{P}_{\mathcal{T}}|}.
\end{equation}
SRA measures the fraction of step pairs in which the essential step receives a higher score than the redundant one.
We report the final SRA by averaging across all examples in the probing dataset.

\subsection{Probing Analysis}
\label{sec:probing_analysis}
We conduct experiments on DeepSeek-R1-Distill-Qwen-1.5B~/~7B and make the following observations; more detailed results are in Appendix~\ref{Details_of_Probe_Results}:
\begin{itemize}[
  leftmargin=0.5em,
  itemsep=0.4pt,
  parsep=1.6pt,
  topsep=0.2pt,
]

\item \textit{During final solution generation, some attention heads selectively focus on essential reasoning steps while ignoring redundant ones, which are mainly located in the middle-to-late layers.}
We refer to these specialized heads as \textbf{Key-Focus Heads}.
As shown in Figure~\ref{fig:KFHs}(a,b), across both evaluated LRMs, a small number of heads achieve SRA above 0.9, with the best-performing head reaching 95\% in the 1.5B model and 96\% in the 7B model.
Most heads, by contrast, perform at or below the random baseline (SRA $\approx 0.5$).

\item \textit{Ensembling step scores $\mathcal{S}_{s_k}$ from multiple attention heads provides only limited gains on the probing dataset.}
In Figure~\ref{fig:KFHs}(c,d), under both \textit{top-$k$ selection by SRA} and \textit{greedy SRA-maximizing selection}, performance improves slightly when a small number of heads are combined, but quickly saturates or degrades as more heads are added.

\item \textit{Under RL training with length penalties, the behavior of KFHs varies little, suggesting that their functional roles are robust.}
We train the model using the reward function described in Eq.~\ref{equ:output-level} and re-run the same probing process on checkpoints at steps 500 and 1000.
The resulting SRA distributions show high consistency across checkpoints, 
with pearson correlation coefficients exceeding 0.85 in all cases.

\item \textit{Even when using non-difficult question in probing, KFHs exhibit a degree of generalization to challenging problems.}
We sample responses on the challenging AIME24 dataset and obtain step scores from the head with the highest SRA.
From each thinking process $\mathcal{T}$ with a correct final answer, we remove either the bottom 30\% (low-scoring) or the top 30\% (high-scoring) steps, and then append the prompt ``\textit{I think I have finished thinking. Now give the final solution step by step.\texttt{</think>}}'' to generate the final solution.
As shown in Figure~\ref{fig:KFHs}(e,f), removing high-scoring steps degrades pass@32, while removing low-scoring steps has little impact.
\end{itemize}

\section{\methodname}
%
%
\begin{figure*}[!htbp] 
\centering  
\includegraphics[width=16cm]{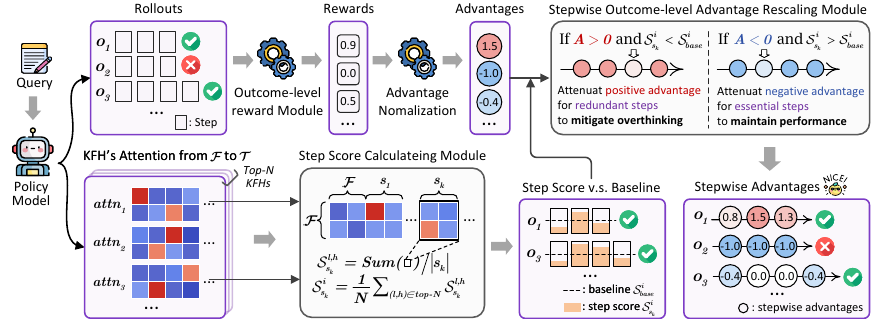}
\caption{The overall framework of \methodname.}
\label{fig:methods}
\vspace{-1em}
\end{figure*}

In this section, we propose \methodname (Attention-guided Policy Optimization), which leverages KFHs to distinguish redundant from essential steps, thereby enabling fine-grained stepwise supervision without additional resources.
We first introduce the overall stepwise advantage rescaling framework in Section~\ref{rescaling}, followed by two complementary strategies under this framework
in Sections~\ref{Pos-Adv} and~\ref{Neg-Adv}.
The overall framework is illustrated in Figure~\ref{fig:methods}.

\subsection{Stepwise Advantage Rescaling}
\label{rescaling}
Current outcome-supervised RL methods for efficient reasoning typically adopt a response-level reward formulation that integrates correctness with length penalties.
The resulting rewards are then leveraged to estimate a value baseline, thereby yielding relative advantages for gradient updates, as exemplified by GRPO~\cite{shao2024deepseekmath} and RLOO~\cite{ahmadian2024back}.
Typically, these algorithms share the following objective function:

\vspace{-1em}
{\small
\begin{multline}
\mathcal{J}(\theta)
=
\underset{\boldsymbol{q} \sim P,\;\{\boldsymbol{o}_i\}_{i=1}^{G} \sim \pi_\theta}{\mathbb{E}}
\Bigg[
\frac{1}{G}
\sum_{i=1}^{G}
\frac{1}{|\boldsymbol{o}_i|}
\sum_{t=1}^{|\boldsymbol{o}_i|}
\Bigg(
\min\!\Big(
\rho_{i,t}\,A^i,\;\\
\operatorname{clip}(\rho_{i,t}, 1-\lambda, 1+\lambda)\,A^i
\Big)
-\beta\, D_{\mathrm{KL}}\!\left(\pi_\theta \,\|\, \pi_{\mathrm{ref}}\right)
\Bigg)
\Bigg].
\end{multline}
}

\noindent Here, $A^i$ denotes the outcome-level advantage, and $G$ the number of rollouts per input $q$.
The importance sampling ratio is
$\rho_{i,t} = \pi_\theta(o_{i,t} \mid q, o_{i,<t}) / \pi_{\theta_{\text{old}}}(o_{i,t} \mid q, o_{i,<t})$.
Although these methods achieve improved performance while eliminating the high-overhead critic model used in PPO~\cite{schulman2017proximal}, they suffer from coarse-grained credit assignment:
the advantage $A^i$ is uniformly applied to all steps within a rollout, regardless of their varying degrees of redundancy.

Thus, we propose \textbf{Stepwise Outcome-level Advantage Rescaling} to rescale the outcome-level advantage $A^i$ at each step $s_k$ of a correct response:
\begin{equation}
\hat{A}^i_{s_k}
= \gamma^{i}_{s_k} \cdot A^i,
 \text{where } \gamma^{i}_{s_k} \ge 0.
\end{equation}
Here, $\hat{A}^i_{s_k}$ denotes the step-level advantage used for policy updates at step $s_k$,
and $\gamma_{s_k}^{i}$ is a non-negative scaling factor computed from the attention scores of KFHs,
reflecting step-level redundancy, which will be detailed in the following section.
This formulation explicitly captures the relative redundancy of individual steps within a single reasoning process.
By scaling the advantage according to stepwise redundancy while preserving its sign,
\methodname enables finer-grained credit assignment without introducing conflicts with the outcome-level supervision.
Notably, advantage scaling is applied only to correct responses,
as identifying essential versus redundant steps in incorrect responses is inherently noisy and can lead to spurious credit assignment.

In this work, we follow the TLMRE~\cite{arora2025training} setup to compute the outcome-level advantage. Specifically, the reward is defined as
\begin{equation}
\label{equ:output-level}
r_{i} = \mathbb{I}[o_i \text{ is correct}]
\left(
1 - \alpha\cdot
\sigma(f(o_i))
\right),
\end{equation}
\begin{equation}
f(o_{i}) = \sigma\big(
(\operatorname{len}(o_{i}) - \operatorname{mean}(q))
/ \operatorname{std}(q)
\big).
\end{equation}

\noindent Here, $|o_{i}|$ is the length of $o_{i}$,
$\operatorname{mean}(q)$ and $\operatorname{std}(q)$ denote the mean and standard deviation of the lengths of correct rollouts sampled by $q$.
$\alpha$ is a hyperparameter.
Following them, we use the RLOO advantage estimator: $A^{i} = r_i - \frac{1}{G-1}\sum_{j \neq i} r_j$.
%
%
%

\subsection{Pos-Adv Attenuation for Redundant-Step}
\label{Pos-Adv}
When a response has a positive outcome-level advantage $A^i$~(Pos-Adv), its generation probability is reinforced. 
Thus, we attenuate advantages for relatively redundant steps to avoid over-encouraging them, thereby mitigating overthinking.
%


First, for each step $s_k$ in response $o_i$, we compute a redundancy-related score $\mathcal{S}_{s_k}^{i}$ by averaging the attention-based step scores (Eq.~\ref{stepscore}) over the top-$N$ KFHs ranked by SRA scores, i.e., $\mathcal{S}_{s_k}^{i} = \frac{1}{k} \sum_{(l,h)\in \text{top-}N} \mathcal{S}_{s_k}^{l,h}$.

Then, we compute a response-specific baseline score $\mathcal{S}_{\text{base}}^{i}$,
which serves as a threshold for distinguishing redundant and essential steps.
Specifically, steps with $\mathcal{S}_{s_k}^{i} < \mathcal{S}_{\text{base}}^{i}$
are treated as redundant, while the remaining steps are considered essential.
The baseline score $\mathcal{S}_{\text{base}}^{i}$ is defined as:
\begin{equation}
\label{equ:baseline}
\mathcal{S}_{\text{base}}^{i}
\triangleq
p_{i}^{\beta}
\cdot
\frac{|\mathcal{F}_i|}{|o_i|},
\text{where }
p_{i}
=
\frac{1}{G}
\sum_{j=1}^{G}
\mathbb{I}[o_j \text{ is correct}].
\end{equation}
Here, $|\mathcal{F}_i| / |o_i|$ measures the average attention received by response tokens
from the final solution $\mathcal{F}_{i}$,
and thus provides a natural reference for identifying relatively under-attended steps as redundant.
%
The term $p_{i}$ serves as a proxy for problem difficulty, estimated by the average rollout accuracy, and $\beta$ is a hyperparameter controlling its influence.
%
The baseline is difficulty-aware, as harder problems demand stricter redundancy criteria due to the increased importance of correctness.




Finally, for correct responses with positive advantages,
we define the stepwise scaling factor $\gamma_{s_k}^{i}$
in a difficulty-aware and scheduled manner:

\vspace{-1em}
\begin{equation}
\small
\label{equ:pos}
\raisetag{1.5ex}
\gamma_{s_k}^{i}
=
\begin{cases}
(1-\delta)\!\cdot\!p_{i}^{\lambda}\!\cdot\!
\left(\dfrac{\mathcal{S}_{s_k}^{i}}{\mathcal{S}_{\text{base}}^{i}}\right)
\!+\!\delta,
&\text{if }\mathcal{S}_{s_k}^{i}\!<\!\mathcal{S}_{\text{base}}^{i} \;\\[-0.7em]
&\land\; t > T \cdot p_{i}, \\
1, & \text{otherwise}.
\end{cases}
\end{equation}

\noindent Here, the ratio $\mathcal{S}_{s_k}^{i} / \mathcal{S}_{\text{base}}^{i}$ serves as a measure of step-level redundancy,
and advantage attenuation is applied only to potentially redundant steps ($\mathcal{S}_{s_k}^{i} < \mathcal{S}_{\text{base}}^{i}$).
The factor $p_{i}^{\lambda}$ controls the attenuation strength with a difficulty-aware magnitude, weakening attenuation for harder problems, where $\lambda$ is a hyperparameter.
$t$ and $T$ denote the current and total training steps, respectively, and the condition $t > T \cdot p_{i}$ defines a difficulty-aware schedule
that delays advantage attenuation on harder problems, by allowing adequate exploration in early stages.
Finally, $\delta$ specifies a lower bound on $\gamma_{s_k}^{i}$.

\subsection{Neg-Adv Attenuation for Essential-Step}
\label{Neg-Adv}
When a correct response receives a negative outcome-level advantage $A^i$ (Neg-Adv), its generation probability is suppressed. 
In this case, we attenuate the penalties on essential steps; as a result, the remaining penalty is relatively focused on redundant ones, thereby avoiding performance degradation caused by penalizing essential reasoning.
Specifically, for correct responses with negative advantages, we define the $\gamma_{s_k}^{i}$ as follows:
\vspace{-0.5em}
\begin{equation}
\gamma_{s_k}^{i}
=
\begin{cases}
0,
&\text{if }\,\mathcal{S}_{s_k}^{i}\!>\!\mathcal{S}_{\text{base}}^{i}, \;\\
1, & \text{otherwise}.
\end{cases}
\end{equation}
Here, we clip negative advantages of potentially essential steps ($\mathcal{S}_{s_k}^{i} > \mathcal{S}_{\text{base}}^{i}$) of correct responses to zero to avoid over-penalization on them.

\begin{table*}[t]
\centering
\setlength{\tabcolsep}{4pt}
\renewcommand{\arraystretch}{1.05}
\resizebox{\textwidth}{!}{%
\begin{tabular}{lccccccccccccccccccccc}
\toprule
\multirow{2}{*}{\raisebox{-0.5\height}{\textbf{Methods}}} & \multicolumn{3}{c}{\textbf{GSM8K}} & \multicolumn{3}{c}{\textbf{MATH500}} & \multicolumn{3}{c}{\textbf{AMC2023}} & \multicolumn{3}{c}{\textbf{Olympiad}} & \multicolumn{3}{c}{\textbf{AIME2024}} & \multicolumn{3}{c}{\textbf{AIME2025}} & \multicolumn{3}{c}{\textbf{Macro Avg}}
\\
\cmidrule(lr){2-4}\cmidrule(lr){5-7}\cmidrule(lr){8-10}\cmidrule(lr){11-13}\cmidrule(lr){14-16}\cmidrule(lr){17-19}\cmidrule(lr){20-22}
 & Acc. & Tok. & AES & Acc. & Tok. & AES & Acc. & Tok. & AES & Acc. & Tok. & AES & Acc. & Tok. & AES & Acc. & Tok. & AES & Acc. & Tok. & AES \\
\midrule
\rowcolor{gray!20} \multicolumn{22}{c}{\textbf{1.5B Model}} \\
DS-1.5B & 78.8 & 1085 & 0 & 82.1 & 5534 & 0 & 65.9 & 7893 & 0 & 49.3 & 9034 & 0 & 28.1 & 12339 & 0 & 22.8 & 12143 & 0 & 54.5 & 8005 & 0 \\
\rowcolor{blue!3} \ding{170}~AutoThink & 83.0 & 568 & 0.64 & 84.0 & 2195 & 0.67 & 67.0 & 5059 & 0.41 & 51.6 & 5053 & 0.58 & 34.6 & 9514 & 0.92 & 21.8 & 7944 & 0.13 & 57.0 & 5056 & 0.56 \\
\ding{170}~AdaptThink & 83.1 & \underline{480} & 0.72 & 82.0 & 1782 & 0.67 & 67.0 & 3510 & 0.61 & 50.8 & 3708 & 0.68 & 31.0 & 6670 & 0.77 & 23.0 & 7255 & 0.43 & 56.1 & 3901 & 0.65 \\
\rowcolor{blue!3} \ding{171}~ThinkPrune & 83.0 & 609 & 0.6 & 83.2 & 1938 & 0.69 & 73.2 & 3039 & 0.95 & 51.4 & 3535 & 0.74 & 27.1 & \underline{5631} & 0.37 & 22.5 & \underline{5167} & 0.51 & 56.7 & \underline{3320} & 0.64 \\
\ding{171}~DIET & - & - & - & 83.0 & 3061 & 0.48 & 65.4 & 6425 & 0.15 & - & - & - & 31.8 & 10578 & 0.54 & - & - & - & - & - & - \\
\rowcolor{blue!3} \ding{171}~ACPO & 81.3 & 572 & 0.57 & 81.0 & \underline{1679} & 0.63 & - & - & - & - & - & - & 30.0 & 6670 & 0.66 & - & - & - & - & - & - \\
\ding{171}~Laser-D & 83.4 & 863 & 0.38 & 84.2 & 1872 & 0.74 & 75.3 & \underline{2981} & 1.05 & 54.4 & 4700 & 0.79 & 34.2 & 5750 & 1.19 & 23.1 & 6928 & 0.47 & 59.1 & 3849 & 0.77 \\
\rowcolor{blue!3} \ding{171}~Laser-DE & 80.4 & 820 & 0.31 & 83.5 & 1949 & 0.7 & 73.3 & 3080 & 0.95 & 54.4 & 5151 & 0.74 & 35.0 & 5789 & 1.27 & 24.2 & 7323 & 0.58 & 58.5 & 4019 & 0.76 \\
\ding{168}~LC-R1 & 82.7 & 841 & 0.37 & 82.5 & 2233 & 0.61 & 61.7 & 3947 & 0.18 & 48.1 & 4546 & 0.38 & 23.6 & 7122 & -0.37 & 21.2 & 6434 & 0.12 & 53.3 & 4187 & 0.22 \\
\rowcolor{blue!3}\ding{168}~VSRM-R++ & - & - & - & 81.7 & 2597 & 0.51 & 64.7 & 4119 & 0.39 & 54.8 & 4388 & 0.85 & 29.5 & 6958 & 0.59 & 22.9 & 6892 & 0.45 & - & - & - \\
 \ding{168}~DEPO & - & - & - & \textbf{87.2} & 2762 & 0.69 & 74.2 & 4388 & 0.82 & - & - & - & 30.8 & 7732 & 0.66 & 24.8 & 7649 & 0.63 & - & - & - \\
\rowcolor{blue!3} \ding{168}~DECS & - & - & - & 84.4 & 1817 & 0.76 & 75.4 & 2988 & 1.05 & 56.1 & \underline{3396} & \underline{1.04} & 31.2 & \textbf{5550} & 0.89 & 23.8 & \textbf{4965} & 0.72 & - & - & - \\
\ding{171}~TLMRE & \textbf{87.2} & 604 & \underline{0.76} & 85.8 & 1915 & \underline{0.79} & \textbf{79.1} & 3349 & \underline{1.18} & \underline{56.9} & 4270 & 0.99 & \underline{37.3} & 7024 & \underline{1.41} & \textbf{26.6} & 7117 & \textbf{0.91} & \textbf{62.2} & 4046 & \underline{1.01} \\
\rowcolor{red!8}  \ding{168}~\methodname & \underline{87.0} & \textbf{393} & \textbf{0.95} & \underline{86.0} & \textbf{1318} & \textbf{0.9} & \underline{77.8} & \textbf{2638} & \textbf{1.21} & \textbf{57.1} & \textbf{3133} & \textbf{1.13} & \textbf{37.7} & 5713 & \textbf{1.56} & \underline{25.4} & 5488 & \underline{0.89} & \underline{61.8} & \textbf{3114} & \textcolor{red}{\textbf{1.11}} \\
\midrule
\rowcolor{gray!20} \multicolumn{22}{c}{\textbf{7B Model}} \\

DS-7B & 88.2 & 639 & 0 & 92.0 & 3593 & 0 & 87.1 & 5977 & 0 & 65.1 & 7358 & 0 & 52.9 & 10490 & 0 & 35.8 & 11307 & 0 & 70.2 & 6561 & 0 \\
\rowcolor{blue!3} \ding{170}~AutoThink & 91.1 & 866 & -0.26 & 91.2 & 2146 & 0.36 & 83.3 & 4645 & 0 & 65.5 & 5133 & 0.32 & 54.8 & 8051 & 0.34 & 36.2 & 8608 & 0.27 & 70.4 & 4908 & 0.17 \\
\ding{170}~AdaptThink & 91.0 & \textbf{309} & \textbf{0.61} & 92.0 & 1875 & 0.48 & 87.5 & 3287 & 0.46 & 65.5 & 5574 & 0.26 & 55.6 & 8599 & 0.33 & 36.2 & 9523 & 0.19 & 71.3 & 4861 & 0.39 \\
\rowcolor{blue!3} \ding{171}~DIET & - & - & - & 92.1 & 3187 & 0.12 & 82.6 & 6075 & -0.27 & - & - & - & 57.9 & 10124 & 0.32 & - & - & - & - & - & - \\
\ding{171}~ACPO & 88.3 & \underline{413} & 0.36 & 91.6 & \underline{1405} & \underline{0.59} & - & - & - & - & - & - & 52.8 & \textbf{4520} & 0.56 & - & - & - & - & - & - \\
\rowcolor{blue!3}\ding{171}~Laser-D & 87.3 & 804 & -0.31 & 92.2 & 1836 & 0.5 & \underline{90.0} & 2694 & \underline{0.65} & 66.7 & 3914 & 0.54 & \underline{58.3} & 5379 & \textbf{0.79} & 38.0 & 6167 & 0.64 & 72.1 & 3466 & 0.47 \\
\ding{171}~Laser-DE & 82.2 & 789 & -0.57 & 92.0 & 1658 & 0.54 & 89.1 & \underline{2612} & 0.63 & 66.8 & 3643 & 0.58 & 55.8 & \underline{4969} & 0.69 & 37.5 & 6077 & 0.61 & 70.6 & \underline{3291} & 0.41 \\

\rowcolor{blue!3} \ding{168}~LC-R1 & 88.1 & 450 & 0.29 & 90.4 & 1568 & 0.48 & 79.1 & 3453 & -0.04 & 64.1 & 4144 & 0.36 & 23.6 & 6904 & -2.42 & 36.2 & 7150 & 0.4 & 63.6 & 3945 & -0.16 \\
\ding{168}~S-GRPO & \textbf{93.8} & 906 & -0.23 & 92.4 & 2252 & 0.39 & 87.5 & 3494 & 0.43 & \underline{69.7} & 3914 & 0.68 & 56.0 & 7377 & 0.47 & 36.0 & 7908 & 0.32 & 72.6 & 4308 & 0.34 \\
\rowcolor{blue!3}\ding{168}~VSRM-R++ & - & - & - & 89.8 & 2044 & 0.31 & 80.9 & 3704 & 0.02 & 66.1 & 5470 & 0.3 & 52.2 & 6773 & 0.29 & 36.4 & 6953 & 0.44 & - & - & - \\
 \ding{168}~DEPO & - & - & - & \textbf{94.4} & 2318 & 0.43 & \textbf{90.5} & 3215 & 0.58 & - & - & - & 52.7 & 6580 & 0.35 & \textbf{39.2} & 7092 & \underline{0.66} & - & - & - \\
\rowcolor{blue!3} \ding{168}~DECS & - & - & - & 93.0 & 1728 & 0.55 & 89.0 & 2772 & 0.6 & \textbf{70.3} & \underline{3283} & \textbf{0.79} & 51.3 & 5277 & 0.35 & 36.4 & \underline{5516} & 0.56 & - & - & - \\
\ding{171}~TLMRE & 91.1 & 515 & 0.29 & \underline{93.8} & 1700 & 0.59 & 89.9 & 2967 & 0.6 & 69.6 & 4454 & 0.6 & \textbf{58.5} & 7462 & 0.61 & \textbf{39.2} & 8301 & 0.55 & \textbf{73.7} & 4233 & \underline{0.54} \\
\rowcolor{red!8}  \ding{168}~\methodname & \underline{92.4} & 446 & \underline{0.44} & 92.8 & \textbf{1300} & \textbf{0.66} & 89.6 & \textbf{2303} & \textbf{0.7} & 68.7 & \textbf{2912} & \underline{0.77} & 57.2 & 5214 & \underline{0.75} & \underline{38.1} & \textbf{5359} & \textbf{0.72} & \underline{73.1} & \textbf{2922} & \textcolor{red}{\textbf{0.67}} \\
\bottomrule
\end{tabular}%
}
\caption{Acc. and Tok. across six benchmarks. Best scores are in \textbf{bold}, second-best are \underline{underlined}. \ding{170}, \ding{171}, and \ding{168} denote adaptive-mode methods, outcome--supervised RL methods and process-supervised RL methods respectively.}
\label{tab:main}
\end{table*}

\section{Experiments}

\subsection{Experiment Setups.}
\paragraph{Evaluation} We evaluate our method on GSM8K~\cite{cobbe2021training}, MATH500~\cite{lightman2023math}, AMC23, OlympiadBench~\cite{he2024olympiadbench}, AIME2024, and AIME2025 as in-domain benchmarks, and on LiveCodeBench-v6~\cite{jain2025livecodebench}, GPQA-Diamond~\cite{rein2024gpqa} and MMLU~\cite{hendrycks2020measuring} as held-out benchmarks, covering math, coding, and science tasks of varying difficulty.
We sample 128 times for AIME2024 and AIME2025, and 4 times for all other benchmarks.
We report Pass@1 (Acc.), response length (Tok.), and compute the Average Efficiency Score (AES)~\cite{luo2025o1} to provide a comprehensive assessment of both effectiveness and efficiency. Evaluation details are provided in Appendix~\ref{sec:eval}.

\paragraph{Training}
We train DeepSeek-R1-Distill-Qwen-1.5B (DS-1.5B) and DeepSeek-R1-Distill-Qwen-7B (DS-7B) on the DeepScaleR-Preview dataset~\cite{deepscaler2025}, using the \textsc{VeRL} framework~\cite{sheng2025hybridflow}.
We set $\beta = 2$ (Eq.~\ref{equ:baseline}), $\lambda = 2$ (Eq.~\ref{equ:pos}), $\delta = 0$ for DS-1.5B and $\delta = 0.5$ for DS-7B, and fix $N = 3$ KFHs.
Additional hyperparameters are reported in Appendix~\ref{sec:train}.

\paragraph{Baselines}
We compare against three categories of methods for efficient reasoning:
(1) Outcome-supervised RL methods, including TLMRE~\cite{arora2025training}, ThinkPrune~\cite{hou2025thinkprune}, DIET~\cite{chen2025overthinker}, ACPO~\cite{cheng2025incentivizing}, and Laser~\cite{liu2025learnreasonefficientlyadaptive};
(2) Process-supervised RL methods, including LC-R1~\cite{cheng2025optimizing}, VSRM-R++~\cite{yue2025promoting}, S-GRPO~\cite{dai2025s}, DEPO~\cite{tan2025towards}, and DECS~\cite{jiang2025overthinking}; and
(3) Adaptive-mode methods, including AdaptThink~\cite{zhang2025adaptthink} and AutoThink~\cite{tu2025learning}, which are reported for reference.
For TLMRE, we re-implement under the same experimental settings as \methodname to ensure fair comparison.
For other methods, we either evaluate their publicly released models or report results from the original papers.

\begin{figure}[!t] 
  \centering
  \includegraphics[width=7.5cm]{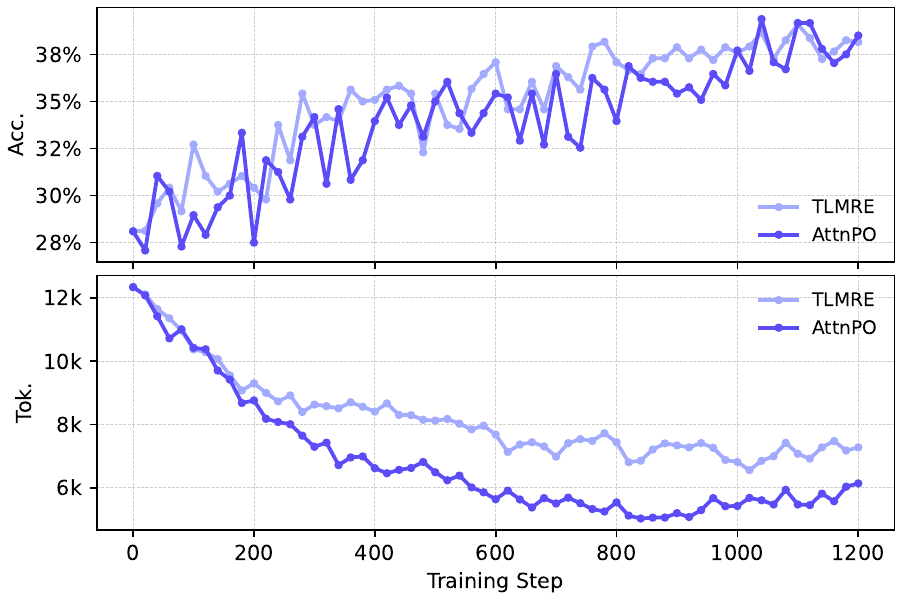}
  \vspace{-0.5em}
  \caption{Training dynamics of TLMRE and \methodname on 1.5B scale on AIME2024 with 16 sampling runs.}
  
  \label{fig:training}
  \vspace{-1em}
\end{figure}

\subsection{Main Results}

\methodname consistently achieves the best or second-best AES across all six math benchmarks and attains the highest average AES across all datasets, indicating a superior efficiency–performance trade-off.
As shown in Table~\ref{tab:main}, \methodname reduces the average length by 61\% on the 1.5B model while improving pass@1 accuracy by +7.3 points.
For example, on the challenging AIME2024 benchmark, \methodname achieves a +9.6 point accuracy gain with a 54\% reduction in length.
On the 7B model, it cuts tokens by 55\% and achieves a +2.9 point accuracy gain, demonstrating efficiency and performance improvements across model scales.
Notably, when compared with TLMRE, the outcome-level advantage used by \methodname, \methodname achieves much larger reductions in length (23\% $\downarrow$ on 1.5B and 31\% $\downarrow$ on  7B) with negligible loss in accuracy (0.4 points $\downarrow$ on 1.5B and 0.6 points $\downarrow$ 7B).
The training dynamics in Figure~\ref{fig:training} on AIME2024 show that \methodname initially underperforms TLMRE on harder tasks due to rapid length compression, but gradually improves and ultimately matches TLMRE in accuracy, indicating that its stepwise advantages enable continued performance gains despite aggressive compression.


\subsection{Ablation Studies}

\begin{figure}[!t] 
  \centering
  \includegraphics[width=7.5cm]{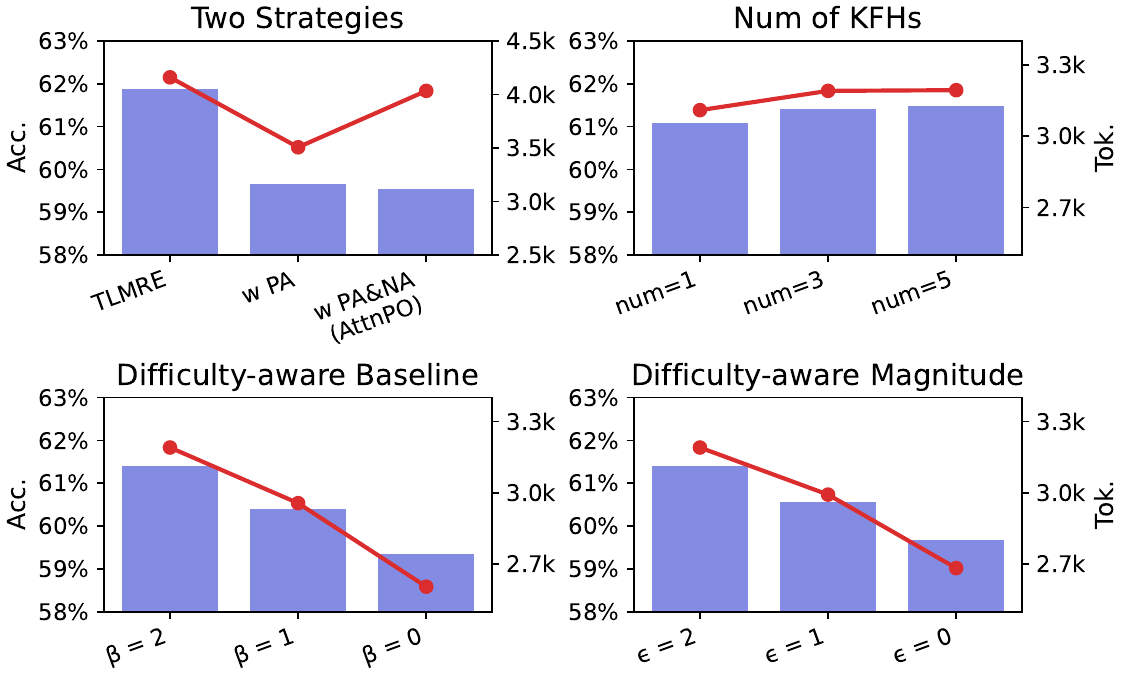}
  \caption{Ablation of strategies \& hyperparameters; \textcolor[HTML]{848be4}{purple bars} show Tok., \textcolor[HTML]{db2d2d}{red line} shows Acc..}
  \label{fig:ablation}
\end{figure}

In this section, we perform ablation studies on 1.5B models to assess different strategies and hyperparameters, with results shown in Figure~\ref{fig:ablation}.

\paragraph{Ablation for two Advantage Rescaling Components.}
Starting from the TLMRE, we sequentially add Positive-Advantage Attenuation (PA) and Negative-Advantage Attenuation (NA).
Applying PA alone substantially reduces reasoning length, indicating that merely weakening the positive advantage of redundant steps provides sufficient optimization pressure to suppress unnecessary reasoning.
Further introducing NA leaves the length almost unchanged while improving accuracy, suggesting that alleviating excessive penalties on necessary reasoning benefits reasoning performance.

\paragraph{Ablation for Num of KFHs.}
Increasing KFHs yields only marginal gains with rapidly diminishing returns, consistent with results in Section~\ref{sec:probing_analysis}, suggesting that a small top-N set of KFHs provides sufficient learning signals.

\paragraph{Ablation for Difficulty-Aware Baselines and Magnitude.}

We ablate the difficulty-aware baseline $\beta$ (Eq.~\ref{equ:baseline}) and magnitude $\lambda$ (Eq.~\ref{equ:pos}) to study their role in the performance–efficiency trade-off.
Weakening ($\beta$ or $\lambda = 1$) and removing (($\beta$ or $\lambda = 0$) the difficulty-based modulation further shortens responses but causes increasing accuracy degradation, indicating that conservative compression on difficult problems is crucial for reducing verbosity while improving performance.
Additionally, we provide an ablation study of the difficulty-aware schedule in Appendix~\ref{app:difficulty-Aware-schedule}.

\begin{figure}[!t] 
  \centering
  \includegraphics[width=6.8cm]{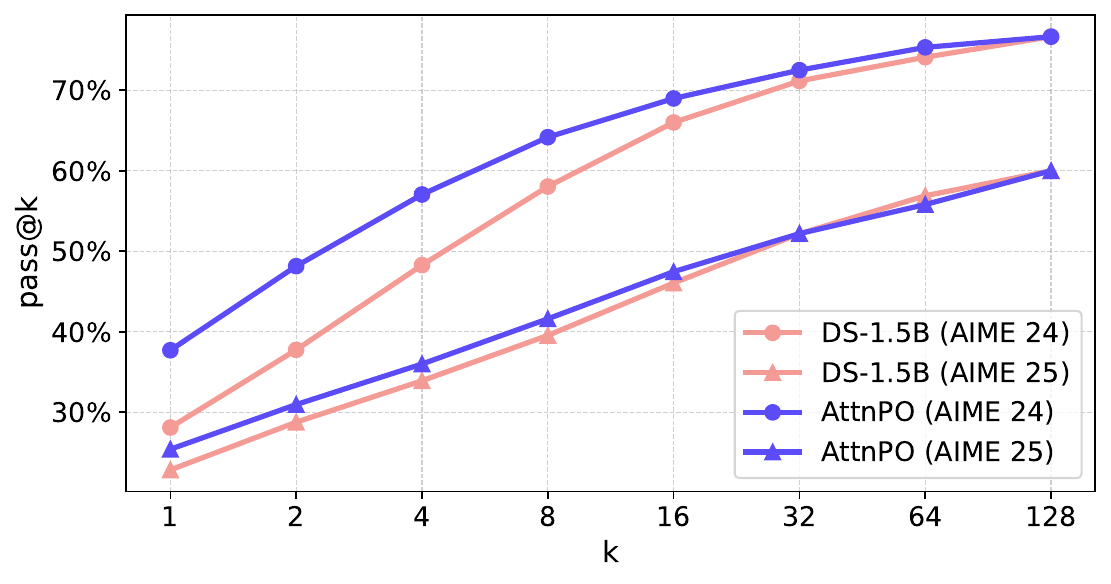}
  \vspace{-1em}
  \caption{Change of pass@k with different k.}
  \label{fig:passk}
\end{figure}

\begin{table}[!t]
\centering
\setlength{\tabcolsep}{4pt}
\resizebox{0.47\textwidth}{!}{
\begin{tabular}{lccccccccc}
\toprule
 & \multicolumn{3}{c}{LiveCodeBench} 
 & \multicolumn{3}{c}{GPQA}
 & \multicolumn{3}{c}{MMLU} \\
\cmidrule(lr){2-4} \cmidrule(lr){5-7} \cmidrule(lr){8-10}
Model & Acc. & Tok. & AES & Acc. & Tok. & AES & Acc. & Tok. & AES \\
\midrule
\rowcolor{gray!20} \multicolumn{10}{c}{\textbf{1.5B Model}} \\
DS-1.5B & 25.3 & 10809 & 0.00 & 33.3 & 8304 & 0.00 & 41.7 & 1872 & 0.00 \\
TLMRE   & 28.6 & 9077  & 0.55 & \textbf{39.2} & 5880 & 0.77 & 48.4 & 1228 & 0.83 \\
\rowcolor{red!8} \methodname  & \textbf{30.4} & \textbf{7567}  & \textbf{0.90} & 36.4 & \textbf{3358} & \textbf{0.87} & \textbf{50.6} & \textbf{603}  & \textbf{1.32} \\
\midrule
\rowcolor{gray!20} \multicolumn{10}{c}{\textbf{7B Model}} \\
DS-7B   & 49.6 & 9095  & 0.00 & 44.6 & 6891 & 0.00 & 69.7 & 1245 & 0.00 \\
TLMRE   & \textbf{53.5} & 3988  & 0.80 & \textbf{49.1} & 3988 & 0.72 & \textbf{70.0} & 712  & 0.44 \\
\rowcolor{red!8} \methodname  & 52.3 & \textbf{2873}  & \textbf{0.85} & 48.6 & \textbf{2873} & \textbf{0.85} & 69.8 & \textbf{547}  & \textbf{0.56} \\
\bottomrule
\end{tabular}
}
\caption{OOD results on LiveCodeBench, GPQA, and MMLU. Best results are shown in \textbf{bold}.}
\label{tab:generalizationl}
\end{table}

\subsection{More Analyses}

To better understand the effects of \methodname, we conduct further analysis on 1.5B models, unless otherwise specified.

\paragraph{Analyses of Exploration Capability.}
To assess whether \methodname’s substantial reduction in reasoning length compromises exploration, we evaluate pass@k ($k\in {2,4,8,16,32,64,128}$) using 128 samples on AIME2024 and AIME2025. As shown in Figure~\ref{fig:passk}, in nearly all sampling budgets, \methodname match or outperform the original model. These results indicate that \methodname achieves significant efficiency gains while retaining much of the model’s exploratory capacity.
%

\paragraph{Analyses of Cross-Domain Capability.}
We analyze the model’s generalization ability on out-of-domain (OOD) tasks. 
In Table~\ref{tab:generalizationl}, \methodname substantially reduces reasoning length on both models, while maintaining or slightly improving accuracy.
These results indicate that, despite being trained only on math, \methodname generalizes well to code and science tasks.

\begin{figure}[!t] 
  \centering
  \includegraphics[width=7.5cm]{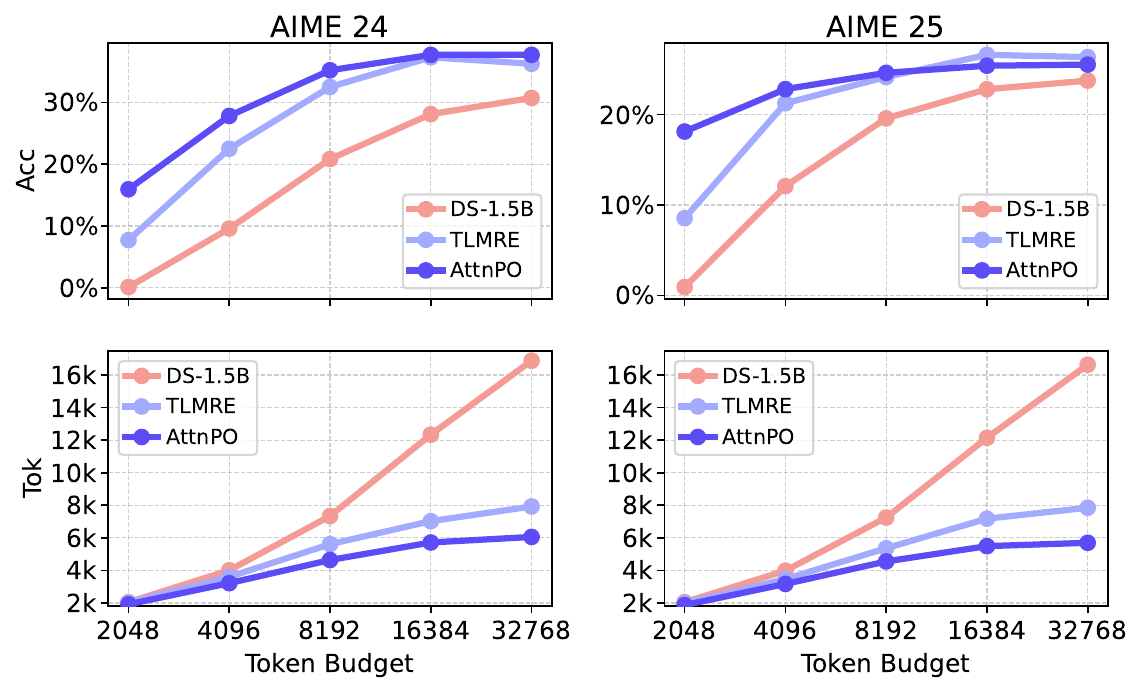}
  \vspace{-0.5em}
  \caption{Acc. and Tok. under different token budgets.}
  \label{fig:budget}
\end{figure}

\paragraph{Analyses of Different Token Budget.}
We evaluate on AIME2024 and AIME2025 using five increasing token budgets (2048, 4096, 8192, 16384, 32768) to examine whether \methodname can improve performance under tight token constraints without sacrificing performance at higher limits. The results show that \methodname significantly outperforms both the base model and TLMRE under tight budgets, and remains superior to the base model while matching TLMRE at higher budgets with substantially shorter outputs, demonstrating the superiority of \methodname across a range of generation length constraints.

\begin{figure}[!t] 
  \centering
  \includegraphics[width=7cm]{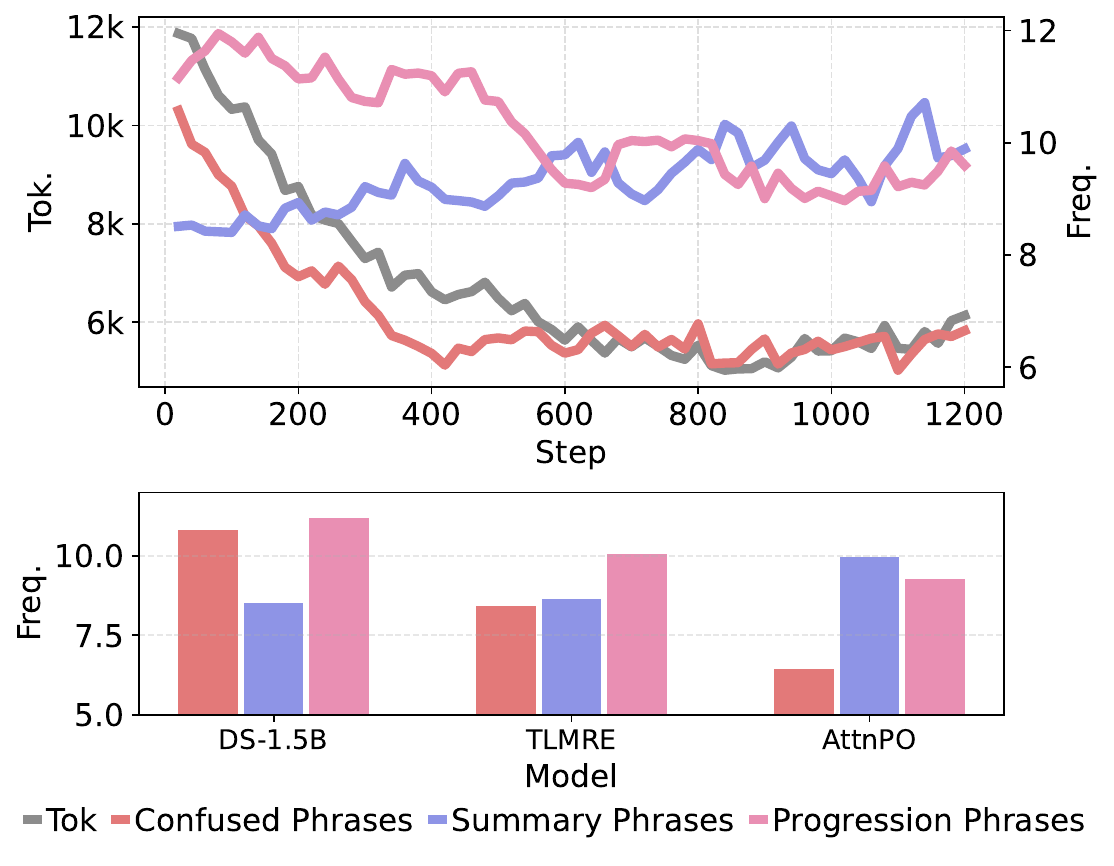}
  \vspace{-0.5em}
  \caption{Changes in special phrases across training steps (top) and models (bottom). Freq. indicates the frequency of special phrases per 1k tokens.}
  \label{fig:evolution}
  \vspace{-0.5em}
\end{figure}

\paragraph{Analyses of Model Behaivors.}
To better understand the response length reduction of \methodname, we analyze the frequency of special phrases with different reasoning behaviors (introduced in section~\ref{sec:Preliminary}) on AIME2024.
we show the evolution of these special phrases during \methodname training (Figure~\ref{fig:evolution}, top) and the final behavioral differences among \methodname, TLMRE, and DS-1.5B (Figure~\ref{fig:evolution}, bottom).
The frequency of Confused Phrases drops sharply during \methodname training, indicating that \methodname effectively reduces redundant self-reflection, this trend is also observed in cross-model comparisons.
The slight increase in Summary Phrases suggests growing confidence in intermediate reasoning results.
Meanwhile, the mild decrease in Progression Phrases implies more consolidated reasoning by skipping trivial intermediate steps.

\section{Related Work}
We first introduce overthinking in LRMs and common mitigation approaches, then focus on outcome-supervised and process-supervised RL methods which most closely related to our work.

\paragraph{Overthinking and Its Mitigation.} 
Benefiting from the RLVR training paradigm~\cite{tan2025scaling,li2026mempo,yu2026knowrl,yang2026soup}, LRMs have achieved strong performance on complex reasoning tasks~\cite{an2025amo}. However, they often exhibit overthinking by generating redundant reasoning~\cite{wei2026evolution,dong2026neu}, which can even degrade performance on certain tasks~\cite{yuan2025understanding}. 
Mitigating overthinking remains a challenging problem. Existing methods mainly fall into five categories: (1) RL with redundancy-based reward design, which we discuss in detail below; (2) curating concise CoT data for SFT or DPO~\cite{yuan2026graph,ma2025cot}; (3) compressing reasoning steps into fewer latent representations~\cite{deng2025latent}; and (4) dynamic reasoning strategy selection~\cite{wang2026adaptive,zhang2025adaptthink}; (5) inference-time intervention methods~\cite{liu2025thought,renze2024benefits,ling2026neural, zhu2026dissecting}.

\paragraph{Outcome-supervised RL method.} 
This line of work typically adopts response-level rewards combining correctness with length-aware penalties. Some approaches explicitly subtract length-correlated penalty terms~\cite{arora2025training, aggarwal2025l1, team2025kimi, yi2025shorterbetter}, while others further refine this by jointly modeling response length and question difficulty~\cite{chen2025overthinker, liu2025learnreasonefficientlyadaptive, xiang2025just}. However, these methods treat the output as a whole and fail to distinguish efficient from inefficient reasoning segments.

\paragraph{Process-supervised RL method.} 
Some methods introduce process supervision to intermediate reasoning steps to distinguish necessary from redundant reasoning.
One line of work incorporates process supervision through additional sampling.
S-GRPO~\cite{dai2025s}, VSRM~\cite{yue2025promoting}, SHAPE~\cite{ai2026shape} and SPAE~\cite{wu2026step} achieve fine-grained credit assignment by allowing early termination at intermediate steps, with the model directly samples the final solution. 
And ROSE~\cite{zhao2026reinforced} extends process supervision to tree-structured rollouts with concise reasoning incentives.
Another line of work introduces auxiliary reward models~\cite{liu2026openrubricsscalablesyntheticrubric}.
DECS~\cite{jiang2025overthinking} and DEPO~\cite{tan2025towards} train an auxiliary to identify when the correct answer first appears and penalize subsequent redundant tokens.
Additionally, since tokens with different entropy may play different roles in reasoning~\cite{wang2025beyond,zhang2026expseek}, some work performs entropy- or confidence-aware advantage shaping~\cite{jin2025revisiting, liu2025uniform, xie2025unlocking}. However, these methods still rely on surface-level entropy signals and often lead to longer reasoning trajectories.
%


\section{Conclusion}

In this paper, we introduce \methodname, a low-overhead process-supervised RL framework that leverages KFHs’ attention signals for stepwise credit assignment. By identifying KFHs that focus on essential steps and designing targeted sub-strategies to suppress redundancy while preserving necessary reasoning, \methodname effectively mitigates overthinking without sacrificing performance.
These results highlight KFH-guided process supervision as an efficient way to improve reasoning efficiency.

\section*{Limitations}

We discuss several limitations of our work in this section.
(1) Owing to limited computational resources, our experiments are conducted only on 1.5B and 7B models. Nonetheless, the results consistently demonstrate the effectiveness of \methodname across different model scales.
(2) In line with most prior open-sourced studies, we train our models exclusively on mathematical datasets, as they are readily available and provide accurate, verifiable reward signals. Although evaluations on LiveCodeBench, GPQA, and MMLU indicate that \methodname generalizes well to OOD scenarios, we anticipate further improvements with the availability of more verifiable training data from broader domains.

\section*{Acknowledgments}
We would like to thank the anonymous reviewers, the meta-reviewer, as well as the area chairs and program chairs for their valuable comments and efforts.
This work is supported by the National Natural Science Foundation of China (Grant No. 62572465).

\bibliography{custom}

@article{guo2025deepseek,
   title={{DeepSeek-R1}: Incentivizing Reasoning Capability in LLMs via Reinforcement Learning},
   volume={645},
   ISSN={1476-4687},
   url={http://dx.doi.org/10.1038/s41586-025-09422-z},
   DOI={10.1038/s41586-025-09422-z},
   number={8081},
   journal={Nature},
   publisher={Springer Science and Business Media LLC},
   author={Guo, Daya and Yang, Dejian and Zhang, Haowei and Song, Junxiao and Wang, Peiyi and Zhu, Qihao and Xu, Runxin and Zhang, Ruoyu and Ma, Shirong and Bi, Xiao and Zhang, Xiaokang and Yu, Xingkai and Wu, Yu and Wu, Z. F. and Gou, Zhibin and Shao, Zhihong and Li, Zhuoshu and Gao, Ziyi and Liu, Aixin and Xue, Bing and Wang, Bingxuan and Wu, Bochao and Feng, Bei and Lu, Chengda and Zhao, Chenggang and Deng, Chengqi and Ruan, Chong and Dai, Damai and Chen, Deli and Ji, Dongjie and Li, Erhang and Lin, Fangyun and Dai, Fucong and Luo, Fuli and Hao, Guangbo and Chen, Guanting and Li, Guowei and Zhang, H. and Xu, Hanwei and Ding, Honghui and Gao, Huazuo and Qu, Hui and Li, Hui and Guo, Jianzhong and Li, Jiashi and Chen, Jingchang and Yuan, Jingyang and Tu, Jinhao and Qiu, Junjie and Li, Junlong and Cai, J. L. and Ni, Jiaqi and Liang, Jian and Chen, Jin and Dong, Kai and Hu, Kai and You, Kaichao and Gao, Kaige and Guan, Kang and Huang, Kexin and Yu, Kuai and Wang, Lean and Zhang, Lecong and Zhao, Liang and Wang, Litong and Zhang, Liyue and Xu, Lei and Xia, Leyi and Zhang, Mingchuan and Zhang, Minghua and Tang, Minghui and Zhou, Mingxu and Li, Meng and Wang, Miaojun and Li, Mingming and Tian, Ning and Huang, Panpan and Zhang, Peng and Wang, Qiancheng and Chen, Qinyu and Du, Qiushi and Ge, Ruiqi and Zhang, Ruisong and Pan, Ruizhe and Wang, Runji and Chen, R. J. and Jin, R. L. and Chen, Ruyi and Lu, Shanghao and Zhou, Shangyan and Chen, Shanhuang and Ye, Shengfeng and Wang, Shiyu and Yu, Shuiping and Zhou, Shunfeng and Pan, Shuting and Li, S. S. and Zhou, Shuang and Wu, Shaoqing and Yun, Tao and Pei, Tian and Sun, Tianyu and Wang, T. and Zeng, Wangding and Liu, Wen and Liang, Wenfeng and Gao, Wenjun and Yu, Wenqin and Zhang, Wentao and Xiao, W. L. and An, Wei and Liu, Xiaodong and Wang, Xiaohan and Chen, Xiaokang and Nie, Xiaotao and Cheng, Xin and Liu, Xin and Xie, Xin and Liu, Xingchao and Yang, Xinyu and Li, Xinyuan and Su, Xuecheng and Lin, Xuheng and Li, X. Q. and Jin, Xiangyue and Shen, Xiaojin and Chen, Xiaosha and Sun, Xiaowen and Wang, Xiaoxiang and Song, Xinnan and Zhou, Xinyi and Wang, Xianzu and Shan, Xinxia and Li, Y. K. and Wang, Y. Q. and Wei, Y. X. and Zhang, Yang and Xu, Yanhong and Li, Yao and Zhao, Yao and Sun, Yaofeng and Wang, Yaohui and Yu, Yi and Zhang, Yichao and Shi, Yifan and Xiong, Yiliang and He, Ying and Piao, Yishi and Wang, Yisong and Tan, Yixuan and Ma, Yiyang and Liu, Yiyuan and Guo, Yongqiang and Ou, Yuan and Wang, Yuduan and Gong, Yue and Zou, Yuheng and He, Yujia and Xiong, Yunfan and Luo, Yuxiang and You, Yuxiang and Liu, Yuxuan and Zhou, Yuyang and Zhu, Y. X. and Huang, Yanping and Li, Yaohui and Zheng, Yi and Zhu, Yuchen and Ma, Yunxian and Tang, Ying and Zha, Yukun and Yan, Yuting and Ren, Z. Z. and Ren, Zehui and Sha, Zhangli and Fu, Zhe and Xu, Zhean and Xie, Zhenda and Zhang, Zhengyan and Hao, Zhewen and Ma, Zhicheng and Yan, Zhigang and Wu, Zhiyu and Gu, Zihui and Zhu, Zijia and Liu, Zijun and Li, Zilin and Xie, Ziwei and Song, Ziyang and Pan, Zizheng and Huang, Zhen and Xu, Zhipeng and Zhang, Zhongyu and Zhang, Zhen},
   year={2025},
   month=sep, pages={633–638} }

@misc{openai-o1,
    title = {Learning to reason with {LLMs}},
    howpublished = {\url{https://openai.com/index/learning-to-reason-with-llms/}},
    author = {{OpenAI}},
    month = {September},
    year = {2024}
}

@misc{schulman2017proximal,
      title={Proximal Policy Optimization Algorithms}, 
      author={John Schulman and Filip Wolski and Prafulla Dhariwal and Alec Radford and Oleg Klimov},
      year={2017},
      eprint={1707.06347},
      archivePrefix={arXiv},
      primaryClass={cs.LG},
      url={https://arxiv.org/abs/1707.06347}, 
}

@misc{shao2024deepseekmath,
      title={DeepSeekMath: Pushing the Limits of Mathematical Reasoning in Open Language Models}, 
      author={Zhihong Shao and Peiyi Wang and Qihao Zhu and Runxin Xu and Junxiao Song and Xiao Bi and Haowei Zhang and Mingchuan Zhang and Y. K. Li and Y. Wu and Daya Guo},
      year={2024},
      eprint={2402.03300},
      archivePrefix={arXiv},
      primaryClass={cs.CL},
      url={https://arxiv.org/abs/2402.03300}, 
}

@inproceedings{ahmadian2024back,
    title = "Back to Basics: Revisiting {REINFORCE}-Style Optimization for Learning from Human Feedback in {LLM}s",
    author = {Ahmadian, Arash  and
      Cremer, Chris  and
      Gall{\'e}, Matthias  and
      Fadaee, Marzieh  and
      Kreutzer, Julia  and
      Pietquin, Olivier  and
      {\"U}st{\"u}n, Ahmet  and
      Hooker, Sara},
    editor = "Ku, Lun-Wei  and
      Martins, Andre  and
      Srikumar, Vivek",
    booktitle = "Proceedings of the 62nd Annual Meeting of the Association for Computational Linguistics (Volume 1: Long Papers)",
    month = aug,
    year = "2024",
    address = "Bangkok, Thailand",
    publisher = "Association for Computational Linguistics",
    url = "https://aclanthology.org/2024.acl-long.662/",
    doi = "10.18653/v1/2024.acl-long.662",
    pages = "12248--12267"
}

@misc{hammoud2025beyond,
      title={Beyond the Last Answer: Your Reasoning Trace Uncovers More than You Think}, 
      author={Hasan Abed Al Kader Hammoud and Hani Itani and Bernard Ghanem},
      year={2025},
      eprint={2504.20708},
      archivePrefix={arXiv},
      primaryClass={cs.CL},
      url={https://arxiv.org/abs/2504.20708}, 
}

@inproceedings{choi2025think,
    title = "Think Clearly: Improving Reasoning via Redundant Token Pruning",
    author = "Choi, Daewon  and
      Lee, Jimin  and
      Tack, Jihoon  and
      Song, Woomin  and
      Dingliwal, Saket  and
      Jayanthi, Sai Muralidhar  and
      Ganesh, Bhavana  and
      Shin, Jinwoo  and
      Galstyan, Aram  and
      Bodapati, Sravan Babu",
    editor = "Christodoulopoulos, Christos  and
      Chakraborty, Tanmoy  and
      Rose, Carolyn  and
      Peng, Violet",
    booktitle = "Findings of the Association for Computational Linguistics: EMNLP 2025",
    month = nov,
    year = "2025",
    address = "Suzhou, China",
    publisher = "Association for Computational Linguistics",
    url = "https://aclanthology.org/2025.findings-emnlp.1169/",
    doi = "10.18653/v1/2025.findings-emnlp.1169",
    pages = "21437--21451",
    ISBN = "979-8-89176-335-7"
}

@misc{chen2024not,
      title={Do NOT Think That Much for 2+3=? On the Overthinking of o1-Like LLMs}, 
      author={Xingyu Chen and Jiahao Xu and Tian Liang and Zhiwei He and Jianhui Pang and Dian Yu and Linfeng Song and Qiuzhi Liu and Mengfei Zhou and Zhuosheng Zhang and Rui Wang and Zhaopeng Tu and Haitao Mi and Dong Yu},
      year={2025},
      eprint={2412.21187},
      archivePrefix={arXiv},
      primaryClass={cs.CL},
      url={https://arxiv.org/abs/2412.21187}, 
}

@inproceedings{vaswani2017attention,
 author = {Vaswani, Ashish and Shazeer, Noam and Parmar, Niki and Uszkoreit, Jakob and Jones, Llion and Gomez, Aidan N and Kaiser, \L ukasz and Polosukhin, Illia},
 booktitle = {Advances in Neural Information Processing Systems},
 editor = {I. Guyon and U. Von Luxburg and S. Bengio and H. Wallach and R. Fergus and S. Vishwanathan and R. Garnett},
 pages = {},
 publisher = {Curran Associates, Inc.},
 title = {Attention is All you Need},
 url = {https://proceedings.neurips.cc/paper_files/paper/2017/file/3f5ee243547dee91fbd053c1c4a845aa-Paper.pdf},
 volume = {30},
 year = {2017}
}

@misc{zheng2024attention,
      title={Attention Heads of Large Language Models: A Survey}, 
      author={Zifan Zheng and Yezhaohui Wang and Yuxin Huang and Shichao Song and Mingchuan Yang and Bo Tang and Feiyu Xiong and Zhiyu Li},
      year={2024},
      eprint={2409.03752},
      archivePrefix={arXiv},
      primaryClass={cs.CL},
      url={https://arxiv.org/abs/2409.03752}, 
}

@misc{deepscaler2025,
  title={DeepScaleR: Surpassing O1-Preview with a 1.5B Model by Scaling RL},
  author={Michael Luo and Sijun Tan and Justin Wong and Xiaoxiang Shi and William Y. Tang and Manan Roongta and Colin Cai and Jeffrey Luo and Li Erran Li and Raluca Ada Popa and Ion Stoica},
  year={2025},
  howpublished={\url{https://pretty-radio-b75.notion.site/DeepScaleR-Surpassing-O1-Preview-with-a-1-5B-Model-by-Scaling-RL-19681902c1468005bed8ca303013a4e2}},
  note={Notion Blog},
}

@misc{maa2025a,
  author = {{Mathematical Association of America}},
  title  = {American Invitational Mathematics Examination 2024},
  year   = {2025}
}

@misc{maa2025b,
  author = {{Mathematical Association of America}},
  title  = {American Invitational Mathematics Examination 2025},
  year   = {2025}
}

@misc{aimo2024,
  author = {{AI-MO}},
  title  = {AMC 2023 Dataset},
  year   = {2024}
}

@inproceedings{he2024olympiadbench,
    title = "{O}lympiad{B}ench: A Challenging Benchmark for Promoting {AGI} with Olympiad-Level Bilingual Multimodal Scientific Problems",
    author = "He, Chaoqun  and
      Luo, Renjie  and
      Bai, Yuzhuo  and
      Hu, Shengding  and
      Thai, Zhen  and
      Shen, Junhao  and
      Hu, Jinyi  and
      Han, Xu  and
      Huang, Yujie  and
      Zhang, Yuxiang  and
      Liu, Jie  and
      Qi, Lei  and
      Liu, Zhiyuan  and
      Sun, Maosong",
    editor = "Ku, Lun-Wei  and
      Martins, Andre  and
      Srikumar, Vivek",
    booktitle = "Proceedings of the 62nd Annual Meeting of the Association for Computational Linguistics (Volume 1: Long Papers)",
    month = aug,
    year = "2024",
    address = "Bangkok, Thailand",
    publisher = "Association for Computational Linguistics",
    url = "https://aclanthology.org/2024.acl-long.211/",
    doi = "10.18653/v1/2024.acl-long.211",
    pages = "3828--3850"
}

@inproceedings{
    lightman2023math,
    title={Let's Verify Step by Step},
    author={Hunter Lightman and Vineet Kosaraju and Yuri Burda and Harrison Edwards and Bowen Baker and Teddy Lee and Jan Leike and John Schulman and Ilya Sutskever and Karl Cobbe},
    booktitle={The Twelfth International Conference on Learning Representations},
    year={2024},
    url={https://openreview.net/forum?id=v8L0pN6EOi}
}

@inproceedings{
    rein2024gpqa,
    title={{GPQA}: A Graduate-Level Google-Proof Q\&A Benchmark},
    author={David Rein and Betty Li Hou and Asa Cooper Stickland and Jackson Petty and Richard Yuanzhe Pang and Julien Dirani and Julian Michael and Samuel R. Bowman},
    booktitle={First Conference on Language Modeling},
    year={2024},
    url={https://openreview.net/forum?id=Ti67584b98}
}

@inproceedings{
    jain2025livecodebench,
    title={LiveCodeBench: Holistic and Contamination Free Evaluation of Large Language Models for Code},
    author={Naman Jain and King Han and Alex Gu and Wen-Ding Li and Fanjia Yan and Tianjun Zhang and Sida Wang and Armando Solar-Lezama and Koushik Sen and Ion Stoica},
    booktitle={The Thirteenth International Conference on Learning Representations},
    year={2025},
    url={https://openreview.net/forum?id=chfJJYC3iL}
}

@inproceedings{
luo2025o1,
title={O1-Pruner: Length-Harmonizing Fine-Tuning for O1-Like Reasoning Pruning},
author={Haotian Luo and Li Shen and Haiying He and Yibo Wang and Shiwei Liu and Wei Li and Naiqiang Tan and Xiaochun Cao and Dacheng Tao},
booktitle={2nd AI for Math Workshop @ ICML 2025},
year={2025},
url={https://openreview.net/forum?id=ioYybCRcyW}
}

@misc{chen2021evaluating,
      title={Evaluating Large Language Models Trained on Code}, 
      author={Mark Chen and Jerry Tworek and Heewoo Jun and Qiming Yuan and Henrique Ponde de Oliveira Pinto and Jared Kaplan and Harri Edwards and Yuri Burda and Nicholas Joseph and Greg Brockman and Alex Ray and Raul Puri and Gretchen Krueger and Michael Petrov and Heidy Khlaaf and Girish Sastry and Pamela Mishkin and Brooke Chan and Scott Gray and Nick Ryder and Mikhail Pavlov and Alethea Power and Lukasz Kaiser and Mohammad Bavarian and Clemens Winter and Philippe Tillet and Felipe Petroski Such and Dave Cummings and Matthias Plappert and Fotios Chantzis and Elizabeth Barnes and Ariel Herbert-Voss and William Hebgen Guss and Alex Nichol and Alex Paino and Nikolas Tezak and Jie Tang and Igor Babuschkin and Suchir Balaji and Shantanu Jain and William Saunders and Christopher Hesse and Andrew N. Carr and Jan Leike and Josh Achiam and Vedant Misra and Evan Morikawa and Alec Radford and Matthew Knight and Miles Brundage and Mira Murati and Katie Mayer and Peter Welinder and Bob McGrew and Dario Amodei and Sam McCandlish and Ilya Sutskever and Wojciech Zaremba},
      year={2021},
      eprint={2107.03374},
      archivePrefix={arXiv},
      primaryClass={cs.LG},
      url={https://arxiv.org/abs/2107.03374}, 
}

@inproceedings{
    arora2025training,
    title={Training Language Models to Reason Efficiently},
    author={Daman Arora and Andrea Zanette},
    booktitle={The Thirty-ninth Annual Conference on Neural Information Processing Systems},
    year={2025},
    url={https://openreview.net/forum?id=AiZxn84Wdo}
}

@article{aggarwal2025l1,
  title={L1: Controlling how long a reasoning model thinks with reinforcement learning},
  author={Aggarwal, Pranjal and Welleck, Sean},
  journal={arXiv preprint arXiv:2503.04697},
  year={2025}
}

@article{team2025kimi,
  publtype={informal},
  author={Kimi Team and Angang Du and Bofei Gao and Bowei Xing and Changjiu Jiang and Cheng Chen and Cheng Li and Chenjun Xiao and Chenzhuang Du and Chonghua Liao and Chuning Tang and Congcong Wang and Dehao Zhang and Enming Yuan and Enzhe Lu and Fengxiang Tang and Flood Sung and Guangda Wei and Guokun Lai and Haiqing Guo and Han Zhu and Hao Ding and Hao Hu and Hao Yang and Hao Zhang and Haotian Yao and Haotian Zhao and Haoyu Lu and Haoze Li and Haozhen Yu and Hongcheng Gao and Huabin Zheng and Huan Yuan and Jia Chen and Jianhang Guo and Jianlin Su and Jianzhou Wang and Jie Zhao and Jin Zhang and Jingyuan Liu and Junjie Yan and Junyan Wu and Lidong Shi and Ling Ye and Longhui Yu and Mengnan Dong and Neo Zhang and Ningchen Ma and Qiwei Pan and Qucheng Gong and Shaowei Liu and Shengling Ma and Shupeng Wei and Sihan Cao and Siying Huang and Tao Jiang and Weihao Gao and Weimin Xiong and Weiran He and Weixiao Huang and Wenhao Wu and Wenyang He and Xianghui Wei and Xianqing Jia and Xingzhe Wu and Xinran Xu and Xinxing Zu and Xinyu Zhou and Xuehai Pan and Y. Charles and Yang Li and Yangyang Hu and Yangyang Liu and Yanru Chen and Yejie Wang and Yibo Liu and Yidao Qin and Yifeng Liu and Ying Yang and Yiping Bao and Yulun Du and Yuxin Wu and Yuzhi Wang and Zaida Zhou and Zhaoji Wang and Zhaowei Li and Zhen Zhu and Zheng Zhang and Zhexu Wang and Zhilin Yang and Zhiqi Huang and Zihao Huang and Ziyao Xu and Zonghan Yang},
  title={Kimi k1.5: Scaling Reinforcement Learning with LLMs},
  year={2025},
  month={January},
  cdate={1735689600000},
  journal={CoRR},
  volume={abs/2501.12599},
  url={https://doi.org/10.48550/arXiv.2501.12599}
}

@article{xiang2025just,
  title={Just Enough Thinking: Efficient Reasoning with Adaptive Length Penalties Reinforcement Learning},
  author={Xiang, Violet and Blagden, Chase and Rafailov, Rafael and Lile, Nathan and Truong, Sang and Finn, Chelsea and Haber, Nick},
  journal={arXiv preprint arXiv:2506.05256},
  year={2025}
}

@article{yi2025shorterbetter,
  title={Shorterbetter: Guiding reasoning models to find optimal inference length for efficient reasoning},
  author={Yi, Jingyang and Wang, Jiazheng and Li, Sida},
  journal={NeurIPS},
  year={2025}
}

@article{chen2025overthinker,
  title={The Overthinker's DIET: Cutting Token Calories with DIfficulty-AwarE Training},
  author={Chen, Weize and Yuan, Jiarui and Jin, Tailin and Ding, Ning and Chen, Huimin and Liu, Zhiyuan and Sun, Maosong},
  journal={NeurIPS},
  year={2025}
}

@misc{liu2025learnreasonefficientlyadaptive,
      title={Learn to Reason Efficiently with Adaptive Length-based Reward Shaping}, 
      author={Wei Liu and Ruochen Zhou and Yiyun Deng and Yuzhen Huang and Junteng Liu and Yuntian Deng and Yizhe Zhang and Junxian He},
      year={2025},
      eprint={2505.15612},
      archivePrefix={arXiv},
      primaryClass={cs.CL},
      url={https://arxiv.org/abs/2505.15612}, 
}

@article{dai2025s,
  title={S-GRPO: Early Exit via Reinforcement Learning in Reasoning Models},
  author={Dai, Muzhi and Yang, Chenxu and Si, Qingyi},
  journal={NeurIPS},
  year={2025}
}

@article{yue2025promoting,
  title={Promoting efficient reasoning with verifiable stepwise reward},
  author={Yue, Chuhuai and Dong, Chengqi and Gao, Yinan and He, Hang and Chai, Jiajun and Yin, Guojun and Lin, Wei},
  journal={arXiv preprint arXiv:2508.10293},
  year={2025}
}

@article{tan2025towards,
  title={Towards Flash Thinking via Decoupled Advantage Policy Optimization},
  author={Tan, Zezhong and Gao, Hang and Ma, Xinhong and Zhang, Feng and Dong, Ziqiang},
  journal={arXiv preprint arXiv:2510.15374},
  year={2025}
}

@article{jiang2025overthinking,
  title={Overthinking Reduction with Decoupled Rewards and Curriculum Data Scheduling},
  author={Jiang, Shuyang and Liao, Yusheng and Zhang, Ya and Wang, Yanfeng and Wang, Yu},
  journal={arXiv preprint arXiv:2509.25827},
  year={2025}
}

@misc{li2025system,
      title={From System 1 to System 2: A Survey of Reasoning Large Language Models}, 
      author={Zhong-Zhi Li and Duzhen Zhang and Ming-Liang Zhang and Jiaxin Zhang and Zengyan Liu and Yuxuan Yao and Haotian Xu and Junhao Zheng and Pei-Jie Wang and Xiuyi Chen and Yingying Zhang and Fei Yin and Jiahua Dong and Zhiwei Li and Bao-Long Bi and Ling-Rui Mei and Junfeng Fang and Xiao Liang and Zhijiang Guo and Le Song and Cheng-Lin Liu},
      year={2025},
      eprint={2502.17419},
      archivePrefix={arXiv},
      primaryClass={cs.AI},
      url={https://arxiv.org/abs/2502.17419}, 
}

@misc{cobbe2021training,
      title={Training Verifiers to Solve Math Word Problems}, 
      author={Karl Cobbe and Vineet Kosaraju and Mohammad Bavarian and Mark Chen and Heewoo Jun and Lukasz Kaiser and Matthias Plappert and Jerry Tworek and Jacob Hilton and Reiichiro Nakano and Christopher Hesse and John Schulman},
      year={2021},
      eprint={2110.14168},
      archivePrefix={arXiv},
      primaryClass={cs.LG},
      url={https://arxiv.org/abs/2110.14168}, 
}

@article{hendrycks2020measuring,
  title={Measuring massive multitask language understanding},
  author={Hendrycks, Dan and Burns, Collin and Basart, Steven and Zou, Andy and Mazeika, Mantas and Song, Dawn and Steinhardt, Jacob},
  journal={arXiv preprint arXiv:2009.03300},
  year={2020}
}

@inproceedings{sheng2025hybridflow,
  title={Hybridflow: A flexible and efficient rlhf framework},
  author={Sheng, Guangming and Zhang, Chi and Ye, Zilingfeng and Wu, Xibin and Zhang, Wang and Zhang, Ru and Peng, Yanghua and Lin, Haibin and Wu, Chuan},
  booktitle={Proceedings of the Twentieth European Conference on Computer Systems},
  pages={1279--1297},
  year={2025}
}

@article{hou2025thinkprune,
  title={Thinkprune: Pruning long chain-of-thought of llms via reinforcement learning},
  author={Hou, Bairu and Zhang, Yang and Ji, Jiabao and Liu, Yujian and Qian, Kaizhi and Andreas, Jacob and Chang, Shiyu},
  journal={arXiv preprint arXiv:2504.01296},
  year={2025}
}

@article{cheng2025incentivizing,
  title={Incentivizing Dual Process Thinking for Efficient Large Language Model Reasoning},
  author={Cheng, Xiaoxue and Li, Junyi and Zhang, Zhenduo and Tang, Xinyu and Zhao, Wayne Xin and Kong, Xinyu and Zhang, Zhiqiang},
  journal={arXiv preprint arXiv:2505.16315},
  year={2025}
}

@article{cheng2025optimizing,
  title={Optimizing Length Compression in Large Reasoning Models},
  author={Cheng, Zhengxiang and Chen, Dongping and Fu, Mingyang and Zhou, Tianyi},
  journal={arXiv preprint arXiv:2506.14755},
  year={2025}
}

@article{tu2025learning,
  title={Learning When to Think: Shaping Adaptive Reasoning in R1-Style Models via Multi-Stage RL},
  author={Tu, Songjun and Lin, Jiahao and Zhang, Qichao and Tian, Xiangyu and Li, Linjing and Lan, Xiangyuan and Zhao, Dongbin},
  journal={arXiv preprint arXiv:2505.10832},
  year={2025}
}

@article{zhang2025adaptthink,
  title={Adaptthink: Reasoning models can learn when to think},
  author={Zhang, Jiajie and Lin, Nianyi and Hou, Lei and Feng, Ling and Li, Juanzi},
  journal={arXiv preprint arXiv:2505.13417},
  year={2025}
}

@misc{zhang2025s1bench,
      title={S1-Bench: A Simple Benchmark for Evaluating System 1 Thinking Capability of Large Reasoning Models}, 
      author={Wenyuan Zhang and Shuaiyi Nie and Xinghua Zhang and Zefeng Zhang and Tingwen Liu},
      year={2025},
      eprint={2504.10368},
      archivePrefix={arXiv},
      primaryClass={cs.CL},
      url={https://arxiv.org/abs/2504.10368}, 
}

@article{ai2026shape,
  title={SHAPE: Stage-aware Hierarchical Advantage via Potential Estimation for LLM Reasoning},
  author={Ai, Zhengyang and Shan, Zikang and Ai, Xiaodong and Tang, Jingxian and Hu, Hangkai and Lu, Pinyan},
  journal={arXiv preprint arXiv:2604.06636},
  year={2026}
}

@article{ling2026neural,
  title={Neural Chain-of-Thought Search: Searching the Optimal Reasoning Path to Enhance Large Language Models},
  author={Ling, Guoming and Huang, Zhongzhan and Lin, Yupei and Li, Junxin and Zhong, Shanshan and Wu, Hefeng and Lin, Liang},
  journal={arXiv preprint arXiv:2601.11340},
  year={2026}
}

@article{an2025amo,
  title={Amo-bench: Large language models still struggle in high school math competitions},
  author={An, Shengnan and Cai, Xunliang and Cao, Xuezhi and Li, Xiaoyu and Lin, Yehao and Liu, Junlin and Lv, Xinxuan and Ma, Dan and Wang, Xuanlin and Wang, Ziwen and others},
  journal={arXiv preprint arXiv:2510.26768},
  year={2025}
}

@article{wu2026step,
  title={Step Potential Advantage Estimation: Harnessing Intermediate Confidence and Correctness for Efficient Mathematical Reasoning},
  author={Wu, Fei and Zhang, Zhenrong and Chang, Qikai and Zhang, Jianshu and Liu, Quan and Du, Jun},
  journal={arXiv preprint arXiv:2601.03823},
  year={2026}
}

@article{wang2025beyond,
  title={Beyond the 80/20 rule: High-entropy minority tokens drive effective reinforcement learning for llm reasoning},
  author={Wang, Shenzhi and Yu, Le and Gao, Chang and Zheng, Chujie and Liu, Shixuan and Lu, Rui and Dang, Kai and Chen, Xionghui and Yang, Jianxin and Zhang, Zhenru and others},
  journal={arXiv preprint arXiv:2506.01939},
  year={2025}
}

@article{jin2025revisiting,
  title={Revisiting Entropy in Reinforcement Learning for Large Reasoning Models},
  author={Jin, Renren and Gao, Pengzhi and Ren, Yuqi and Han, Zhuowen and Zhang, Tongxuan and Huang, Wuwei and Liu, Wei and Luan, Jian and Xiong, Deyi},
  journal={arXiv preprint arXiv:2511.05993},
  year={2025}
}

@article{liu2025uniform,
  title={From Uniform to Heterogeneous: Tailoring Policy Optimization to Every Token's Nature},
  author={Liu, Zheng and Liu, Mengjie and Wen, Siwei and Cai, Mengzhang and Cui, Bin and He, Conghui and Zhang, Wentao},
  journal={arXiv preprint arXiv:2509.16591},
  year={2025}
}

@article{xie2025unlocking,
  title={Unlocking exploration in rlvr: Uncertainty-aware advantage shaping for deeper reasoning},
  author={Xie, Can and Pan, Ruotong and Wu, Xiangyu and Zhang, Yunfei and Fu, Jiayi and Gao, Tingting and Zhou, Guorui},
  journal={arXiv preprint arXiv:2510.10649},
  year={2025}
}

@article{wei2026evolution,
  title={The Evolution of Thought: Tracking LLM Overthinking via Reasoning Dynamics Analysis},
  author={Zihao Wei and Liang Pang and Jiahao Liu and Wenjie Shi and Jingcheng Deng and Shicheng Xu and Zenghao Duan and Fei Sun and Huawei Shen and Xueqi Cheng},
  journal={arXiv preprint arXiv:2508.17627},
  year={2025}
}

@article{yuan2025understanding,
  title={Understanding LLM Reasoning for Abstractive Summarization},
  author={Yuan, Haohan and Zhang, Haopeng},
  journal={arXiv preprint arXiv:2512.03503},
  year={2025}
}

@article{tan2025scaling,
  title={Scaling behaviors of llm reinforcement learning post-training: An empirical study in mathematical reasoning},
  author={Tan, Zelin and Geng, Hejia and Yu, Xiaohang and Zhang, Mulei and Wan, Guancheng and Zhou, Yifan and He, Qiang and Xue, Xiangyuan and Zhou, Heng and Fan, Yutao and others},
  journal={arXiv preprint arXiv:2509.25300},
  year={2025}
}

@article{li2025cai,
  title={CAI: Caption-Sensitive Attention Intervention for Mitigating Object Hallucination in Large Vision-Language Models},
  author={Li, Qiming and Ye, Zekai and Feng, Xiaocheng and Zhong, Weihong and Qin, Libo and Chen, Ruihan and Li, Baohang and Jiang, Kui and Wang, Yaowei and Liu, Ting and others},
  journal={arXiv preprint arXiv:2506.23590},
  year={2025}
}

@article{chen2026sparse,
  title={Sparse Growing Transformer: Training-Time Sparse Depth Allocation via Progressive Attention Looping},
  author={Chen, Yao and Chen, Yilong and Yang, Yinqi and Shang, Junyuan and Zhang, Zhenyu and Zhang, Zefeng and Nie, Shuaiyi and Wang, Shuohuan and Sun, Yu and Wu, Hua and others},
  journal={arXiv preprint arXiv:2603.23998},
  year={2026}
}

@inproceedings{ma2025cot,
  title={Cot-valve: Length-compressible chain-of-thought tuning},
  author={Ma, Xinyin and Wan, Guangnian and Yu, Runpeng and Fang, Gongfan and Wang, Xinchao},
  booktitle={Proceedings of the 63rd Annual Meeting of the Association for Computational Linguistics (Volume 1: Long Papers)},
  pages={6025--6035},
  year={2025}
}

@article{deng2025latent,
  title={Latent reasoning in llms as a vocabulary-space superposition},
  author={Deng, Jingcheng and Pang, Liang and Wei, Zihao and Xu, Shichen and Duan, Zenghao and Xu, Kun and Song, Yang and Shen, Huawei and Cheng, Xueqi},
  journal={arXiv preprint arXiv:2510.15522},
  year={2025}
}

@inproceedings{wang2026adaptive,
  title={Adaptive Social Learning via Mode Policy Optimization for Language Agents},
  author={Wang, Minzheng and Li, Yongbin and Wang, Haobo and Zhang, Xinghua and Xu, Nan and Wu, Bingli and Huang, Fei and Yu, Haiyang and Mao, Wenji},
  booktitle={The Fourteenth International Conference on Learning Representations},
  year={2026}
}

@article{zhao2026reinforced,
  title={Reinforced Efficient Reasoning via Semantically Diverse Exploration},
  author={Zhao, Ziqi and Ren, Zhaochun and Zou, Jiahong and Yang, Liu and Xu, Zhiwei and Ge, Xuri and Chen, Zhumin and Ma, Xinyu and Shi, Daiting and Wang, Shuaiqiang and others},
  journal={arXiv preprint arXiv:2601.05053},
  year={2026}
}

@article{yuan2026graph,
  title={Graph-Based Chain-of-Thought Pruning for Reducing Redundant Reflections in Reasoning LLMs},
  author={Yuan, Hongyuan and He, Xinran and Shao, Run and He, Bolei and Xue, Xianwei and Chen, Mengke and Pan, Qiutong and Wang, Haiwei and Li, Haifeng},
  journal={arXiv preprint arXiv:2604.05643},
  year={2026}
}

@misc{liu2026openrubricsscalablesyntheticrubric,
      title={OpenRubrics: Towards Scalable Synthetic Rubric Generation for Reward Modeling and LLM Alignment}, 
      author={Tianci Liu and Ran Xu and Tony Yu and Ilgee Hong and Carl Yang and Tuo Zhao and Haoyu Wang},
      year={2026},
      eprint={2510.07743},
      archivePrefix={arXiv},
      primaryClass={cs.CL},
      url={https://arxiv.org/abs/2510.07743}, 
}

@article{zhang2026expseek,
  title={ExpSeek: Self-Triggered Experience Seeking for Web Agents},
  author={Zhang, Wenyuan and Zhang, Xinghua and Yu, Haiyang and Nie, Shuaiyi and Wu, Bingli and Yue, Juwei and Liu, Tingwen and Li, Yongbin},
  journal={arXiv preprint arXiv:2601.08605},
  year={2026}
}

@article{li2026mempo,
  title={MemPO: Self-Memory Policy Optimization for Long-Horizon Agents},
  author={Li, Ruoran and Zhang, Xinghua and Yu, Haiyang and Duan, Shitong and Li, Xiang and Xiang, Wenxin and Liao, Chonghua and Guo, Xudong and Li, Yongbin and Suo, Jinli},
  journal={arXiv preprint arXiv:2603.00680},
  year={2026}
}

@article{liu2025thought,
  title={Thought manipulation: External thought can be efficient for large reasoning models},
  author={Liu, Yule and Zheng, Jingyi and Sun, Zhen and Peng, Zifan and Dong, Wenhan and Sha, Zeyang and Cui, Shiwen and Wang, Weiqiang and He, Xinlei},
  journal={arXiv preprint arXiv:2504.13626},
  year={2025}
}

@article{renze2024benefits,
  title={The benefits of a concise chain of thought on problem-solving in large language models},
  author={Renze, Matthew and Guven, Erhan},
  journal={arXiv preprint arXiv:2401.05618},
  year={2024}
}

@misc{yu2026knowrl,
      title={KnowRL: Boosting LLM Reasoning via Reinforcement Learning with Minimal-Sufficient Knowledge Guidance}, 
      author={Linhao Yu and Tianmeng Yang and Siyu Ding and Renren Jin and Naibin Gu and Xiangzhao Hao and Shuaiyi Nie and Deyi Xiong and Weichong Yin and Yu Sun and Hua Wu},
      year={2026},
      eprint={2604.12627},
      archivePrefix={arXiv},
      primaryClass={cs.AI},
      url={https://arxiv.org/abs/2604.12627}, 
}

@article{yang2026soup,
  title={SOUP: Token-level Single-sample Mix-policy Reinforcement Learning for Large Language Models},
  author={Yang, Lei and Bi, Wei and Sun, Chenxi and Jin, Renren and Xiong, Deyi},
  journal={arXiv preprint arXiv:2601.21476},
  year={2026}
}

@article{jiang2026foe,
  title={FoE: Forest of Errors Makes the First Solution the Best in Large Reasoning Models},
  author={Jiang, Kehan and Dong, Haonan and Kang, Zhaolu and Zhu, Zhengzhou and Song, Guojie},
  journal={arXiv preprint arXiv:2604.02967},
  year={2026}
}

@misc{dong2026neu,
      title={NeuReasoner: Towards Explainable, Controllable, and Unified Reasoning via Mixture-of-Neurons}, 
      author={Haonan Dong and Kehan Jiang and Haoran Ye and Wenhao Zhu and Zhaolu Kang and Guojie Song},
      year={2026},
      eprint={2604.02972},
      archivePrefix={arXiv},
      primaryClass={cs.CL},
      url={https://arxiv.org/abs/2604.02972}, 
}

@misc{zhu2026dissecting,
      title={Dissecting Failure Dynamics in Large Language Model Reasoning}, 
      author={Wei Zhu and Jian Zhang and Lixing Yu and Kun Yue and Zhiwen Tang},
      year={2026},
      eprint={2604.14528},
      archivePrefix={arXiv},
      primaryClass={cs.AI},
      url={https://arxiv.org/abs/2604.14528}, 
}

\appendix
\section{Appendix}
\label{sec:appendix}

\subsection{Step Segmentation Algorithm}
\label{sec:Step_Segmentation_Algorithm}

The thinking process $\mathcal{T}$ can be decomposed into a sequence of meaningful intermediate steps, whose boundaries are often signaled by specific special phrases. These phrases mark key moments in the reasoning process, such as pauses, reflections, shifts in direction, or transitions to new stages~\cite{hammoud2025beyond,choi2025think}. Accordingly, we segment the thinking process using a predefined set of special phrases, categorized into \textbf{confusion phrases} (e.g., `Wait''), which indicate reflection, verification, or branching; \textbf{progression phrases} (e.g., First''), which advance the ongoing reasoning; and \textbf{summary phrases} (e.g., Therefore''), which consolidate preceding content. The full set is listed in Table~\ref{tab:special_words}. 

In LRMs, a double newline (\texttt{\textbackslash n\textbackslash n}) typically signals a natural semantic transition~\cite{chen2024not}. To avoid over-segmentation, we treat a special phrases as a segmentation boundary only when it is preceded by a double newline, except for confusion phrases. Since confusion phrases often trigger redundant verification or reflection, we do not impose the newline constraint on segments starting with them, keeping such segments separate from the preceding context to facilitate redundancy detection and penalization.

After initial segmentation, we apply adaptive merging to short segments to avoid semantically weak fragments, with the merging direction determined by the category of the initiating special phrases.
Segments introduced by confusion phrases are always merged with the right neighbor to preserve their separation from prior reasoning.
Segments introduced by progression phrases are merged based on context: if its right neighbor is a confusion-initiated segment, it is merged with the left neighbor to preserve the independence of the confusion segment; otherwise, it is merged with the right neighbor, as progression phases typically advance the reasoning flow.
Segments introduced by summary phrases are merged with the left neighbor, as they serve to consolidate preceding reasoning.

Throughout segmentation and merging, we keep segments starting with confusion phrases separate from the preceding context, as they often correspond to redundant verification or reflection. Isolating such segments facilitates the identification and penalization of redundant reasoning.

We provide the pseudocode in Algorithm~\ref{alg:cot_segmentation}.

\begin{algorithm}[t!]
\footnotesize
\caption{Step Segmentation Algorithm}
\label{alg:cot_segmentation}
\begin{algorithmic}[1]
\Input A thinking process $\mathcal{T}$, 3 predefined special-phrase sets 
$\mathcal{W}_{\mathrm{conf}}$, $\mathcal{W}_{\mathrm{prog}}$, and $\mathcal{W}_{\mathrm{summ}}$, 
and a minimum length threshold $t$ for each step
\Output Segmented steps $[s_1, s_2, \dots, s_K]$

\Statex \textbf{Step 1: Initial Segmentation}
\State $\mathcal{B} \gets [\ ]$ \Comment{boundary positions (in left-to-right order)}
\State $\mathcal{P} \gets [\ ]$ \Comment{segmentation phrase at each boundary}
\State $\mathcal{W} \gets \mathcal{W}_{\mathrm{conf}} \cup \mathcal{W}_{\mathrm{prog}} \cup \mathcal{W}_{\mathrm{summ}}$
\State sort phrases in $\mathcal{W}$ by decreasing length and compile a regex matcher $\mathsf{R}$
\ForAll{matches $(p,w)$ found by $\mathsf{R}$ when scanning $\mathcal{T}$ left-to-right}
  \Comment{$p$: start position of the match; $w$: matched phrase}
  \If{$w \in \mathcal{W}_{\mathrm{prog}}\cup \mathcal{W}_{\mathrm{summ}}$ \textbf{and} $w$ is not preceded by \texttt{\textbackslash n\textbackslash n}}
    \State \textbf{continue}
  \EndIf
  \State append $p$ to $\mathcal{B}$; append $w$ to $\mathcal{P}$
\EndFor

\State if $\mathcal{B}[0]\neq 0$, prepend $0$ to $\mathcal{B}$ and $\textsc{None}$ to $\mathcal{P}$
\State if $\mathcal{B}[|\mathcal{B}|-1]\neq|\mathcal{T}|$, append $|\mathcal{T}|$ to $\mathcal{B}$

\For{$i = 1$ to $|\mathcal{B}|-1$}
  \State $s_i \gets \mathcal{T}\big[\mathcal{B}[i] : \mathcal{B}[i+1]\big]$
\EndFor
\State $K \gets |\mathcal{B}|-1$ \Comment{total number of initial segmented steps}

\Statex \textbf{Step 2: Adaptive Merging}

\Function{MergeRight}{$i$} 
  \State $s_i \gets s_i \circ s_{i+1}$; delete $s_{i+1}$
  \State delete $\mathcal{P}[i+1]$; update $K$ = $K - 1$
\EndFunction

\Function{MergeLeft}{$i$} 
  \State $s_{i-1} \gets s_{i-1} \circ s_i$; delete $s_i$
  \State delete $\mathcal{P}[i]$; update $K$ = $K - 1$
\EndFunction

\State $i \gets 0$
\While{$i < K$}
  \If{$\textsc{len}(s_i) < t$}
    \If{$\mathcal{P}[i] \in \mathcal{W}_{\mathrm{conf}}$} \Comment{confusion}
      \If{$i+1< K$}
        \State \Call{MergeRight}{$i$}
        \State \textbf{continue}
      \EndIf

    \ElsIf{$\mathcal{P}[i] \in \mathcal{W}_{\mathrm{summ}}$} \Comment{summary}
        \If{$i-1 >= 0$}
            \State \Call{MergeLeft}{$i$}
            \State $i \gets i-1$
            \State \textbf{continue}
        \EndIf

      \Else \Comment{progression}
      \If{$i + 1 < K$ \textbf{and} $\mathcal{P}[i+1] \in \mathcal{W}_{\mathrm{conf}}$}
        \If{$i-1 >= 0$}
          \State \Call{MergeLeft}{$i$}
          \State $i \gets i-1$
          \State \textbf{continue}
        \EndIf
      \Else 
        \If{$i +1  < K$}
          \State \Call{MergeRight}{$i$}
          \State \textbf{continue}
        \EndIf
      \EndIf

    \EndIf
  \EndIf
  \State $i \gets i+1$
\EndWhile

\State \Return $[s_1, s_2, \dots, s_K]$
\end{algorithmic}
\end{algorithm}

\begin{table}[htbp]
\centering

\begin{tabularx}{\linewidth}{X}
\toprule
\small\textbf{Confusion phrases:}\\
\small
``\textit{Wait}'', ``\textit{But}'', ``\textit{However}'', ``\textit{Hold on}'', ``\textit{Looking back}'',
``\textit{I don't see any errors}'',
``\textit{Hmm, wait}'', ``\textit{Hmm, no}'', ``\textit{Hmm, but}'', ``\textit{Hmm, just}'',
``\textit{Hmm, let me double-check}'', ``\textit{Hmm, let me verify}'',
``\textit{Hmm, let me make sure}'', ``\textit{Hmm, let me check}'', ``\textit{No,}'',
``\textit{Let me double-check}'', ``\textit{Let me just double-check}'',
``\textit{Let's double-check}'', ``\textit{Let's just double-check}'',
``\textit{Just to double-check}'',
``\textit{Let me verify}'', ``\textit{Let me just verify}'',
``\textit{Let's verify}'', ``\textit{Let's just verify}'',
``\textit{Just to verify}'',
``\textit{Let me confirm}'', ``\textit{Let me just confirm}'',
``\textit{Let's confirm}'', ``\textit{Let's just confirm}'',
``\textit{Just to confirm}'',
``\textit{Let me check}'', ``\textit{Let me just check}'',
``\textit{Let's check}'', ``\textit{Let's just check}'',
``\textit{Just to check}'',
``\textit{Let me recap}'', ``\textit{Let me just recap}'',
``\textit{Let's recap}'', ``\textit{Let's just recap}'',
``\textit{Just to recap}'',
``\textit{Let me make sure}'', ``\textit{Let me just make sure}'',
``\textit{Let's make sure}'', ``\textit{Just to make sure}'',
``\textit{So, is}'', ``\textit{Is it}'', ``\textit{Is that right?}'',
``\textit{Is there}'', ``\textit{Right?}'',
``\textit{Alternatively}'', ``\textit{Another approach}'', ``\textit{Another way}'',
``\textit{Another idea}'', ``\textit{Another thought}'',
``\textit{I guess another way}'', ``\textit{I guess another approach}'',
``\textit{Let me just think if there}'' \\
\midrule
\small\textbf{Progression phrases:}\\
\small
``\textit{Let me}'', ``\textit{Let's}'', ``\textit{I }'', ``\textit{We }'',
``\textit{Okay}'', ``\textit{Hmm}'', ``\textit{Now}'', ``\textit{Alright}'',
``\textit{First,}'', ``\textit{First off,}'', ``\textit{Second,}'', ``\textit{Third,}'',
``\textit{Starting}'', ``\textit{Then}'', ``\textit{Next}'', ``\textit{Finally}'',
``\textit{Similarly}'', ``\textit{Again}'',
``\textit{In this case}'', ``\textit{Because}'',
``\textit{Given that}'', ``\textit{The problem gives}'',
``\textit{To }'', ``\textit{Since}'' \\
\midrule
\small\textbf{Summary phrases:}\\
\small
``\textit{So}'', ``\textit{Therefore}'', ``\textit{Thus}'', ``\textit{Hence}'',
``\textit{Hmm, so}'', ``\textit{Okay, so}'',
``\textit{Putting it all together}''\\
\bottomrule
\end{tabularx}
\caption{Special phrases used for step segmentation.}
\label{tab:special_words}
\end{table}
\subsection{Details of the Probing Experiment}
\subsubsection{Details of Probe Data Construction}
\label{sec:Probe_Data_Construction}
In this subsection, we describe the complete process for constructing the probing dataset.
We select 300 questions from Deepscaler~\cite{deepscaler2025} that are not difficult for DeepSeek-R1-Distill-Qwen-1.5B, on which the model achieves an average accuracy of at least 0.875 over eight sampled responses. This restriction is important because, on more challenging questions, LRMs tend to engage in exploratory reasoning. Although such reasoning may appear redundant, it can still be necessary for reaching correct solutions, making it difficult to reliably distinguish redundant steps from essential ones and potentially confounding the identification of Key-Focus Heads.
For each selected question, we choose the correct response with median length among the eight samples and segment its thinking process into steps following the procedure described in Section~\ref{sec:Step_Segmentation_Algorithm}. 
Each reasoning step is then annotated by gemini-2.5-pro, gpt-5-2025-08-07, claude-sonnet-4-20250514 with the prompt provide in Table~\ref{prompt:redundant-necessary}, and only steps with unanimous agreement are retained to reduce annotation noise. 
The resulting inter-annotator agreement, measured by Krippendorff’s $\alpha$, is 0.64 for DeepSeek-R1-Distill-Qwen-7B and 0.59 for DeepSeek-R1-Distill-Qwen-1.5B, indicating moderate agreement.
We further retain only trajectories containing both essential and redundant steps, so that each probe instance provides an internal essential–redundant contrast, enabling attention comparisons within the same context.

\subsubsection{Details of Probe Results}
\label{Details_of_Probe_Results}
\begin{figure*}[ht] 
\centering  
\includegraphics[width=16cm]{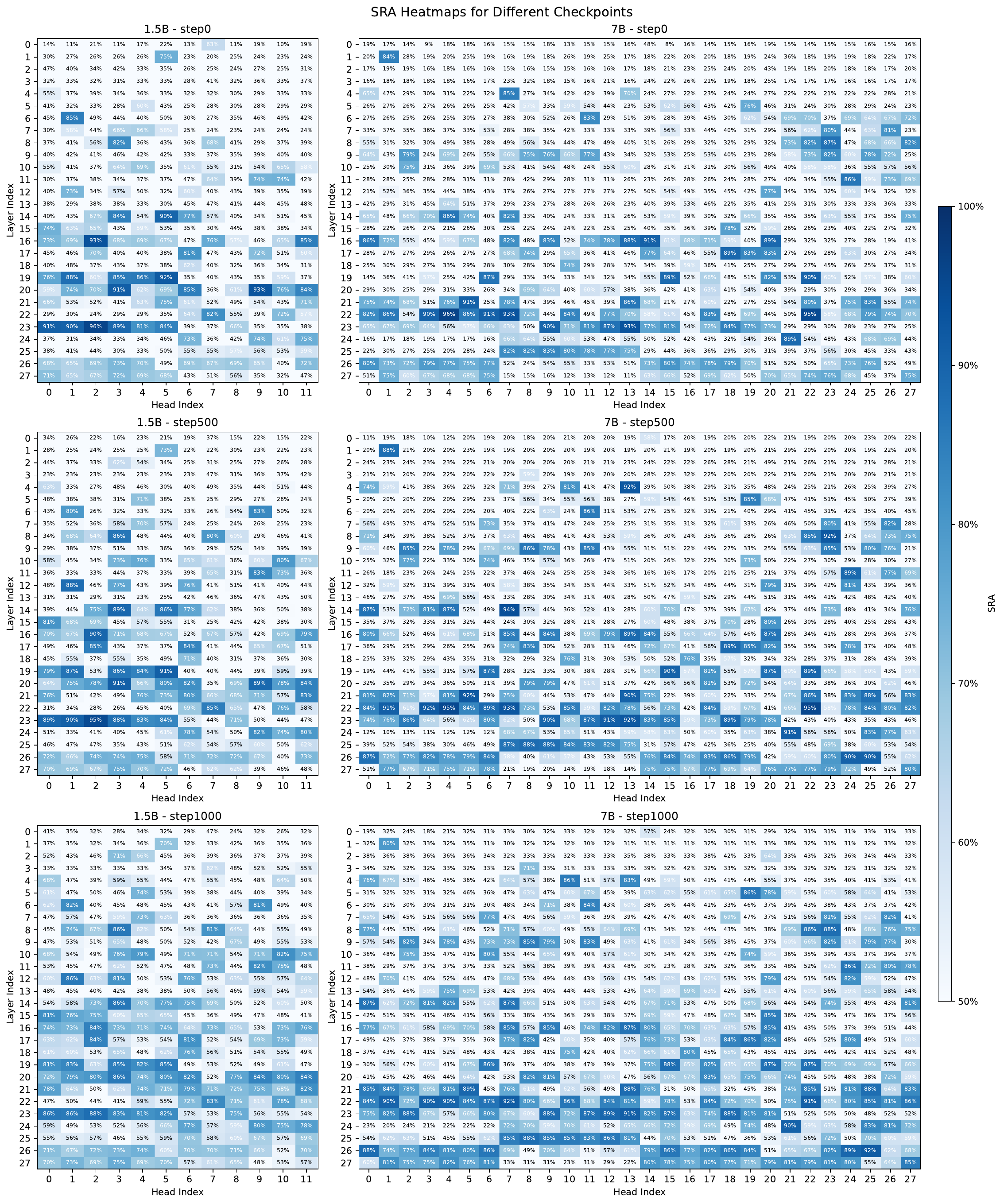}
\caption{SRA heatmap for all models.}
\label{fig:app_all_model_sra}
\end{figure*}

Here, we present the SRA heatmap across the checkpoints at 0, 500, and 1000 steps in Figure~\ref{fig:app_all_model_sra}, and list the Pearson correlation coefficients of SRA among these checkpoints in Table~\ref{tab:step_comparison}.
Additionally, in section~\ref{sec:probing_analysis}, \textit{Top-k selection by SRA} refers to ranking all attention heads by their average SRA scores and selecting the top k heads with the highest scores.
\textit{Greedy SRA-maximizing selection} refers to iteratively selecting attention heads by evaluating all remaining candidates at each step and greedily adding the head that yields the largest increase in the overall SRA score.

\begin{table}[htbp]
    \centering
    \small
    \begin{tabular}{lcc}
        \toprule
        \textbf{Model} & 0 vs 500 & 0 vs 1000 \\
        \midrule
        DS-R1-1.5B & 0.92 & 0.88 \\
        DS-R1-7B & 0.93 &  0.89 \\
        \bottomrule
    \end{tabular}
    \caption{Pearson correlation coefficients of SRA between checkpoints at 0, 500, and 1000 steps.}
    \label{tab:step_comparison}
\end{table}

\section{Experimental Details}
\subsection{Training Hyperparameters}
\label{sec:train}
For the output-level reward in Eq.~\ref{equ:output-level}, we set $\alpha = 0.2$ for the 1.5B model and $\alpha = 0.1$ for the 7B model. This choice is based on our empirical observation that using $\alpha = 0.2$ for the 7B model leads to overly aggressive length compression and severe performance degradation, even without applying our method.
For the KFHs used in both models, we select heads 16-2, 20-9, and 23-2 for the 1.5B model, and heads 22-4, 22-7, and 22-22 for the 7B model.
For the 1.5B model, we train using a single 8×H800 node, while for the 7B model, training is conducted on four 8×H800 nodes.
We present the other hyperparameters adopted during training in Table~\ref{tab:train_param}.

\begin{table}[!htbp]
\centering
\small
\setlength{\tabcolsep}{4pt}
\begin{tabular}{lc}
\toprule
\textbf{Param}  \\
\midrule
max response length        & 16384  \\
batch szie          & 128  \\
rollout top-p  & 1.0   \\
rollout temperature & 1.0 \\
learning rate             & 2e-6  \\
kl loss Coef & 0   \\
rollout number      & 16    \\
$\epsilon_{\text{low}}$     & 0.2  \\
$\epsilon_{\text{high}}$     & 0.2  \\
steps         & 1200    \\
\bottomrule
\end{tabular}
\caption{Training hyperparameters.}
\label{tab:train_param}
\end{table}




\subsection{Evaluation}
\label{sec:eval}

\subsubsection{Descriptions of Benchmarks}

\begin{enumerate}[itemsep=2pt, topsep=4pt]
    \item \textbf{AIME2024, AIME2025} \cite{maa2025a, maa2025b}. 
    These two datasets consist of high-school Olympiad-level assessments from the American Invitational Mathematics Examination (AIME) held in 2024 and 2025, respectively. 
    Each dataset contains 30 highly challenging problems spanning algebra, geometry, and number theory.

    \item \textbf{AMC23} \cite{aimo2024}. 
    This dataset is sourced from the American Mathematics Competitions (AMC) system in 2023 and includes 40 problems with mixed and hybrid question formats.

    \item \textbf{OlympiadBench} \cite{he2024olympiadbench}. 
    This benchmark comprises a comprehensive collection of mathematical Olympiad problems from multiple countries. 
    We select only the English-language math subset and retain problems that require numerical answers, resulting in a total of 581 evaluation problems.

    \item \textbf{MATH500} \cite{lightman2023math}. 
    This dataset is an advanced mathematics evaluation benchmark curated by OpenAI, containing 500 problems expressed with formal mathematical notation.

    \item \textbf{GPQA-Diamond} \cite{rein2024gpqa}. 
    This dataset is a curated subset of the GPQA (Graduate-Level Google-Proof Q\&A) benchmark and consists of 198 challenging multiple-choice questions authored and verified by domain experts in biology, physics, and chemistry.

    \item \textbf{LiveCodeBench} \cite{jain2025livecodebench}. 
    This benchmark is designed to evaluate the live code generation capabilities of large language models, emphasizing immediate correctness and practical programming skills. 
    We use version v6 of the dataset, which contains 1,055 problems in total.

    \item \textbf{MMLU} \cite{hendrycks2020measuring}.
    MMLU is a massive multitask benchmark of multiple-choice questions spanning 57 subjects, including elementary mathematics, U.S. history, computer science, and law. Achieving high accuracy requires extensive world knowledge and strong problem-solving ability. We sample 50 questions from each category for evaluation.
\end{enumerate}

\subsubsection{Evaluation Setups}
We sample 128 times for AIME2024 and AIME2025, and 4 times for all other benchmarks.
We set the evaluation context size to 16K and the temperature to 0.6, as suggested in DeepSeek’s model cards.

\subsubsection{Computation of Metrics}
\label{sec:computation_metrics}

\paragraph{AES}
The AES score~\citep{luo2025o1} is calculated by jointly considering the pass@1 performance and the average token cost of the tuned policy relative to the base policy:
\begin{equation}
\small
\label{aes}
\mathrm{AES}
=
\frac{L_{\rm base} - L}{L_{\rm base}}
+
\left\{
\begin{array}{@{}l@{}}
3 \cdot \dfrac{\mathrm{pass\text{@}1}-\mathrm{pass\text{@}1}_{\rm base}}{\mathrm{pass\text{@}1}_{\rm base}}, \\[4pt]
if~\mathrm{pass\text{@}1} \ge \mathrm{pass\text{@}1}_{\rm base} \\[6pt]
-5 \cdot \dfrac{\mathrm{pass\text{@}1}_{\rm base}-\mathrm{pass\text{@}1}}{\mathrm{pass\text{@}1}_{\rm base}}, \\[4pt]
if~\mathrm{pass\text{@}1} < \mathrm{pass\text{@}1}_{\rm base}
\end{array}
\right.
\end{equation}
This metric captures both the relative reduction in token usage and its effect on model performance: it penalizes approaches that reduce performance and rewards those that yield improvements over the baseline.

\paragraph{Pass@k}
The pass@k metric~\citep{chen2021evaluating} is computed as follows:
\begin{equation}
    \mathrm{pass\text{@}k}=1 - \frac{ \binom{n - c}{k} }{ \binom{n}{k} }
\end{equation}
where $n$ denotes the total number of samples and $c$ denotes the number of correct samples.
When $k=1$, this metric reduces to the average accuracy across the $n$ samples.

\begin{table*}[t!]
\centering
\small
\setlength{\tabcolsep}{4pt}
\renewcommand{\arraystretch}{1.05}
\begin{tabular}{lcccccccccccccc}
\toprule
\multirow{2}{*}{\raisebox{-0.5\height}{\textbf{Methods}}}
& \multicolumn{2}{c}{GSM8K} 
& \multicolumn{2}{c}{MATH500} 
& \multicolumn{2}{c}{AMC2023} 
& \multicolumn{2}{c}{Olympiad} 
& \multicolumn{2}{c}{AIME2024} 
& \multicolumn{2}{c}{AIME2025} 
& \multicolumn{2}{c}{Avg} \\
\cmidrule(lr){2-3}\cmidrule(lr){4-5}\cmidrule(lr){6-7}\cmidrule(lr){8-9}\cmidrule(lr){10-11}\cmidrule(lr){11-12}\cmidrule(lr){12-13} \cmidrule(lr){14-15}
 & Acc. & Tok. & Acc. & Tok. & Acc. & Tok.  & Acc. & Tok. & Acc. & Tok. & Acc. & Tok. & Acc. & Tok. \\
\midrule
TMLER 
& 87.2 & 604 
& 85.8 & 1915 
& 79.1 & 3349 
& 56.9 & 4270 
& 37.3 & 7024 
& 26.6 & 7117 
& 62.2 & 4047 \\

w PA 
& 85.9 & 380 
& 84.4 & 1295 
& 76.8 & 2558 
& 54.2 & 3240 
& 36.8 & 5860 
& 25.0 & 5670 
& 60.5 & 3167 \\

w PA\&NA (\methodname) 
& 87.0 & 393 
& 86.0 & 1318 
& 77.8 & 2638 
& 57.1 & 3133 
& 37.7 & 5713 
& 25.4 & 5488 
& 61.8 & 3114 \\
\bottomrule
\end{tabular}
\caption{Ablation results for two sub-strategies, PA and NA.}
\label{tab:app-two-strategies}
\end{table*}

\section{Additional Experimental Results}

\subsection{Ablation for two sub-strategies}
We present the ablation results for the two sub-strategies in Tables~\ref{tab:app-two-strategies}.

\subsection{Time Overhead Analysis of \methodname}
\begin{figure}[t] 
  \centering
  \includegraphics[width=7.5cm]{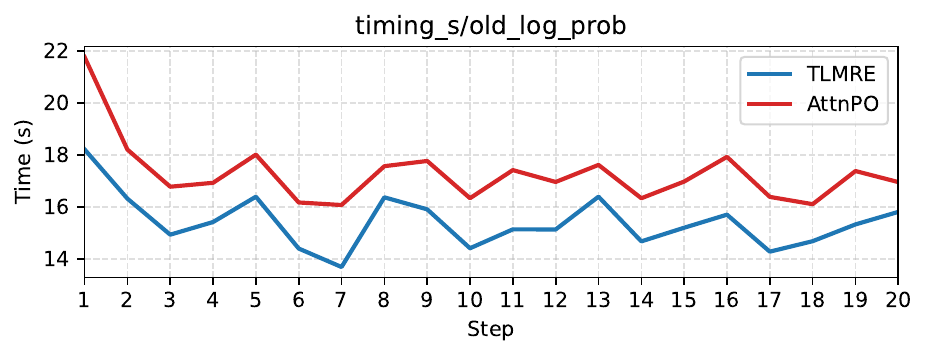}
  \caption{Time Comparison of \methodname and TLMRE on the 1.5B model.}
  \label{fig:time}
\end{figure}
We analyze the additional GPU time overhead introduced by \methodname, which mainly stems from computing attention-based scores for each reasoning step using KFH heads. 
Since our implementation is built on the \textsc{VERL} framework, this step is integrated into the computation of the log-probabilities of the rollout policy (referred to as \texttt{old\_log\_prob} in \textsc{VERL}).
Specifically, during the model’s forward pass for computing importance sampling terms (e.g., log-probabilities and related quantities), we simultaneously compute the attention-based scores for each reasoning step, thereby avoiding additional forward passes and minimizing extra computational overhead.
We report the training time of the 1.5B model under \methodname and TLMRE for the first 20 training steps as a comparison. We focus on the initial 20 steps because the generated response lengths under both methods are nearly identical in this stage, which effectively eliminates the confounding impact of length differences on the measured training time.
The results in Figure~\ref{fig:time} demonstrate that \methodname introduces almost no additional training time overhead.

\subsection{Ablation for Difficulty-Aware Schedule}
\label{app:difficulty-Aware-schedule}
\begin{figure}[t] 
  \centering
  \includegraphics[width=7.5cm]{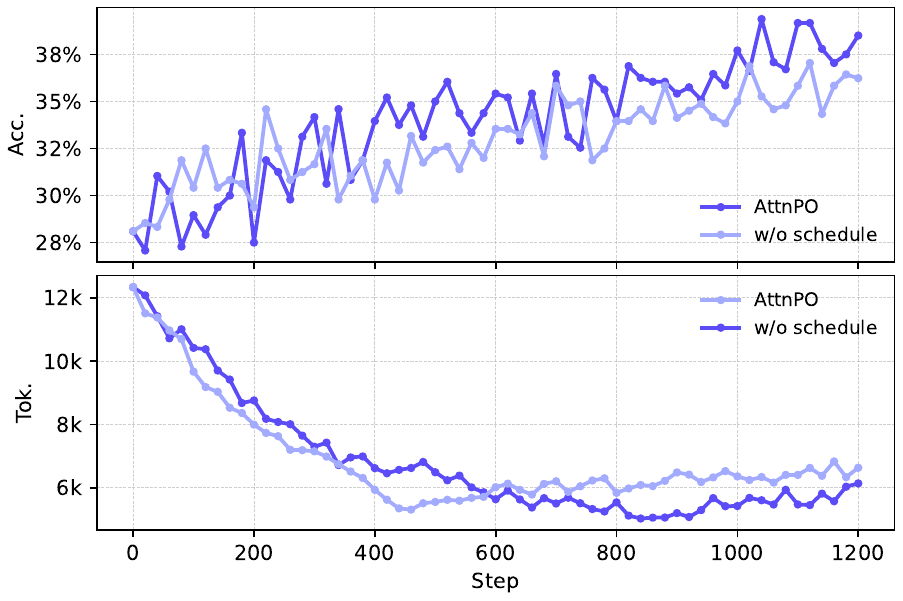}
  \caption{Ablation Results on the Difficulty-Aware Schedule on AIME2024 with 16 sampling runs.}
  \label{fig:app-training}
\end{figure}

As shown in the training dynamics in Figure~\ref{fig:app-training}, introducing a difficulty-aware schedule results in a slower reduction of reasoning length during the early stages of training, but sustains a more consistent compression trend throughout training. In contrast, removing the difficulty-aware schedule leads to a faster initial decrease in reasoning length, followed by a clear plateau in later stages.
This behavior suggests that delaying advantage attenuation on difficult problems allows the model to prioritize accuracy improvement in the early phase, preventing premature and excessive compression. Once performance is sufficiently improved, the model gains greater flexibility to further reduce reasoning length, leading to a more favorable performance–efficiency trade-off.

\subsection{Case Study}
We provide case studies of TLMRE and \methodname, with details presented in Table~\ref{case1},~\ref{case2},~\ref{case3} and~\ref{case4}.
\begin{table*}[!t]
\small
\centering
\begin{tabular}{@{}p{\linewidth}@{}}
\toprule
\textbf{Prompts for Redundant and Necessary Step Annotation} \\
\midrule

\parbox[t]{\linewidth}{\RaggedRight
\textbf{Role \& Task:}\\
You are an expert evaluator of step-by-step reasoning quality for large language models.\\
I will give you a question and a corresponding model response (the response contains explicit step markers, e.g., \texttt{<STA\_2>} means step 2 starts, \texttt{<END\_2>} means step 2 ends). The final result in this response is correct.
Please classify each step as: redundant / necessary / uncertain.\\
If a step contains only redundant content, label it as redundant; only necessary content, label it as necessary; if it contains both redundant and necessary content, label it as uncertain.

\textbf{Category Definitions:}\\
\vspace{0.4em}
\textbf{(1) Redundant:} Steps that do not substantially advance the solution, or repeated/ineffective actions taken after the correct conclusion has already been reached, including but not limited to:
\vspace{-0.6em}
\begin{itemize}\setlength\itemsep{-0.2em}
  \item \textbf{Redundant verification:} When the current reasoning is already correct, but the model performs an additional verification step. This is usually labeled as redundant. Only when the second verification is used to discover and correct an error, or to add essential justification, should it be labeled as necessary.
  \item \textbf{Redundant reflection:} When the current reasoning is already correct, but the model adds extra reflection or self-questioning that merely loops over the same reasoning without introducing new information, or even introduces incorrect information or baseless speculation. These are usually labeled as redundant. Only when such reflection is used to discover and correct an error, or to add essential justification, should it be labeled as necessary.
  \item \textbf{Redundant branches:} After arriving at the correct answer through valid reasoning, the model switches to another method to re-derive the correct answer. This is usually labeled as redundant.
  \item \textbf{Erroneous reasoning:} If a step contains incorrect statements and does not bring any effective progress, it should be considered redundant.
\end{itemize}
\vspace{-0.6em}
\textbf{(2) Necessary:} Steps that positively contribute to advancing the reasoning process, including but not limited to:
\vspace{-0.6em}
\begin{itemize}\setlength\itemsep{-0.2em}
  \item \textbf{Normal computation/reasoning:} Necessary problem analysis, reasoning plans, formalization of the reasoning, intermediate calculations, fact retrieval, and explicit statements of conclusions that help push the reasoning forward should be labeled as necessary.
  \item \textbf{Necessary verification/reflection/branches:} When there are errors, contradictions, dead ends, or uncertainty due to multiple possible solutions, verification, reflection, or branching that is used to correct the direction, eliminate errors, or fill in missing conditions should be labeled as necessary.
\end{itemize}
\vspace{-0.6em}
\textbf{(3) Uncertain:}
\vspace{-0.6em}
\begin{itemize}\setlength\itemsep{-0.2em}
  \item \textbf{Containing both redundant and necessary content:} If a single step includes both redundant content and necessary content, label it as uncertain.
  \item \textbf{Reasonable exploration:} If a step is not on the main line of reasoning but is relevant, correct, and introduces new information, label it as uncertain.
  \item \textbf{Borderline cases:} For borderline cases where you cannot clearly determine whether a step is redundant or necessary, label it as uncertain.
\end{itemize}
\vspace{-0.6em}
\textbf{(4) Borderline Cases:}
\vspace{-0.6em}
\begin{itemize}\setlength\itemsep{-0.2em}
  \item Core planning used to guide subsequent reasoning steps is usually considered necessary.
  \item The model’s final explicit statement of the final answer at the end of the reasoning process is generally considered necessary.
\end{itemize}
\vspace{-0.6em}
\textbf{Question and Model Response:}\\
\vspace{0.2em}
\textbf{Question:}\\
\texttt{\{question\}}\\
\vspace{0.2em}
\textbf{Model response:}\\
\texttt{\{response\}}\\
\vspace{0.6em}
\textbf{Output Requirements:}\\
First, output the reasoning for your judgment on each step, in the following format:\\
\texttt{Step index: reasoning for the judgment}\\
Then output the classification result for each step, in the following format:\\
\texttt{Step index: category (use numbers, 1 = necessary, 2 = redundant, 3 = uncertain; for example, 0:1 means step 0 is necessary)}\\

\vspace{0.4em}
\textbf{Example output (strictly follow this format):}\\
\texttt{[Reasoning for each step]}\\
\texttt{0: Reasoning for the judgment...}\\
\texttt{1: Reasoning for the judgment...}\\
\texttt{2: Reasoning for the judgment...}\\
\texttt{[Classification results]}\\
\texttt{0:1}\\
\texttt{1:2}\\
\texttt{2:3}\\

\vspace{0.4em}
\textbf{Begin your answer:}
} \\
\bottomrule
\end{tabular}

\caption{Prompts for Redundant and Necessary Step Annotation.}
\label{prompt:redundant-necessary}
\end{table*}

\begin{figure*}
\centering
\small
\begin{tcolorbox}[size=title,opacityfill=0.1]
\noindent
\faUser\ \textbf{Question (MATH500)}: If $f(x) =\frac{3x-2}{x-2}$, what is the value of $f(-2) +f(-1)+f(0)$? Express your answer as a common fraction. \\

\faCogs\ \textbf{TLMRE}:
Okay, so I need to find the value of $f(-2)+f(-1)+f(0)$ where $f(x)$ is given by $(3x-2)/(x-2)$. Hmm, let me break this down step by step. First, let me compute $f(-2)$. I substitute $x=-2$ into the function: $f(-2)=(3\cdot(-2)-2)/(-2-2)=(-6-2)/(-4)=(-8)/(-4)=2$. Alright, $f(-2)$ is 2. Next, $f(-1)$. Substitute $x=-1$: $f(-1)=(3\cdot(-1)-2)/(-1-2)=(-3-2)/(-3)=(-5)/(-3)=5/3$. So, $f(-1)$ is $5/3$. Now, $f(0)$. Substitute $x=0$: $f(0)=(3\cdot0-2)/(0-2)=(-2)/(-2)=1$. So, $f(0)$ is 1. Now, add them all together: $f(-2)+f(-1)+f(0)=2+5/3+1$. Convert all to thirds: 2 is $6/3$, 1 is $3/3$, so: $6/3+5/3+3/3=(6+5+3)/3=14/3$. \textcolor{red}{Wait, that seems correct? Let me double-check each calculation. $f(-2)$: $(3\cdot(-2)-2)=-6-2=-8$; denominator $(-2-2)=-4$; $-8/-4=2$. Correct. $f(-1)$: $(3\cdot(-1)-2)=-3-2=-5$; denominator $(-1-2)=-3$; $-5/-3=5/3$. Correct. $f(0)$: $(0-2)/(-2)=-2/-2=1$. Correct. Sum: $2+5/3+1$. $2+1=3$, $3+5/3=3+1.666\ldots=4.666\ldots$, which is $14/3$. Yeah, $14/3$ is correct as a common fraction.} \textbf{Final Answer} The value of $(f(-2)+f(-1)+f(0))$ is $\boxed{\dfrac{14}{3}}$.
\\

{\color[HTML]{2EA121}\faCogs\ \textbf{\textsc{AttnPO}-1.5B (ours)}}: \\
Okay, so I have this function $f(x)=(3x-2)/(x-2)$, and I need to find $f(-2)+f(-1)+f(0)$. Hmm, let's take it step by step. First, let's find $f(-2)$. Plugging $x=-2$ into the function: $f(-2)=(3\cdot(-2)-2)/(-2-2)=(-6-2)/(-4)=(-8)/(-4)=2$. Okay, that's straightforward. Next, $f(-1)$. Plugging $x=-1$: $f(-1)=(3\cdot(-1)-2)/(-1-2)=(-3-2)/(-3)=(-5)/(-3)=5/3$. So that's $5/3$. Then, $f(0)$. Plugging $x=0$: $f(0)=(3\cdot0-2)/(0-2)=(-2)/(-2)=1$. That's simple too. Now, adding them up: $f(-2)+f(-1)+f(0)=2+5/3+1$. Let's convert everything to thirds: $2=6/3$, $1=3/3$. So $6/3+5/3+3/3=(6+5+3)/3=14/3$. \textbf{Final Answer} $\boxed{\dfrac{14}{3}}$.

\end{tcolorbox}

\caption{This is a simple problem from MATH500, where an unnecessarily reflective reasoning process is applied by TLMRE-1.5B. In contrast, the AttnPO-1.5B provides a concise solution.}
\label{case1}
\end{figure*}

\begin{figure*}
\centering
\small
\begin{tcolorbox}[size=title,opacityfill=0.1]
\noindent
\faUser\ \textbf{Question (AIME 2024)}: Every morning Aya goes for a $9$-kilometer-long walk and stops at a coffee shop afterwards. When she walks at a constant speed of $s$ kilometers per hour, the walk takes her 4 hours, including $t$ minutes spent in the coffee shop. When she walks $s+2$ kilometers per hour, the walk takes her 2 hours and 24 minutes, including $t$ minutes spent in the coffee shop. Suppose Aya walks at $s+\frac{1}{2}$ kilometers per hour. Find the number of minutes the walk takes her, including the $t$ minutes spent in the coffee shop. \\

\faCogs\ \textbf{TLMRE}:
Okay, let's try to figure out this problem step by step. Aya walks 9 kilometers every morning and stops at a coffee shop. The time she takes depends on her walking speed. We need to find how long her walk will take if she walks at a speed of $s+\frac{1}{2}$ km/h, including the time spent in the coffee shop. First, let's parse the information given. When she walks at speed $s$ km/h, the total time taken, including the coffee shop time, is 4 hours. When her speed increases to $s+2$ km/h, the total time becomes 2 hours and 24 minutes, which is $2+\frac{24}{60}=2.4$ hours. \textcolor{red}{We need to find the time it takes when she walks at $s+\frac{1}{2}$ km/h. Let me denote: Let $t$ be the time spent in the coffee shop in minutes. Let $d$ be the distance she walks before entering the coffee shop, which is 9 km minus the coffee shop time converted to distance. Wait, no. Actually, the total distance is 9 km, but the coffee shop time is $t$ minutes, but the walking time is different. Wait, maybe I need to model this differently.} Let me denote: Let walking speed be $s$ km/h. Let $t$ be the time spent in the coffee shop in hours. So, when she walks at $s$ km/h, the total time is walking time + $t=4$ hours. Walking time is $9/s$ hours. So, $9/s+t=4$. Similarly, when her speed is $s+2$ km/h, the total time is 2 hours and 24 minutes, which is $2+\frac{24}{60}=2.4$ hours. So, the walking time is $9/(s+2)$ hours, and the total time is that plus $t$, which equals 2.4 hours. Therefore, we have two equations: 1. $9/s+t=4$ 2. $9/(s+2)+t=2.4$ We can solve these two equations to find $s$ and $t$. Subtracting equation 2 from equation 1: $(9/s+t)-(9/(s+2)+t)=4-2.4$. Simplify: $9/s-9/(s+2)=1.6$. Factor out 9: $9\left[1/s-1/(s+2)\right]=1.6$. Compute $1/s-1/(s+2)=(s+2-s)/[s(s+2)]=2/[s(s+2)]$. So, $9\times[2/(s(s+2))]=1.6$. Simplify: $18/[s(s+2)]=1.6$. Multiply both sides by $s(s+2)$: $18=1.6s(s+2)$. Divide both sides by 1.6: $18/1.6=s(s+2)$. Calculate $18/1.6=11.25$. So, $s(s+2)=11.25$. Which is: $s^2+2s-11.25=0$. Quadratic equation: $s^2+2s-11.25=0$. We can solve for $s$ using quadratic formula: $s=[-2\pm\sqrt{4+45}]/2$. Because discriminant is $2^2-4\cdot1\cdot(-11.25)=4+45=49$. So, $s=[-2\pm7]/2$. Since speed can't be negative, take the positive solution: $s=(-2+7)/2=5/2=2.5$ km/h. So, $s=2.5$ km/h. Now, we can find $t$ from equation 1: $9/s+t=4$. $9/2.5+t=4$. $9/2.5$ is 3.6. So, $3.6+t=4$. Therefore, $t=4-3.6=0.4$ hours. Convert 0.4 hours to minutes: $0.4\times60=24$ minutes. So, $t=24$ minutes. Now, when she walks at $s+\frac{1}{2}$ km/h, which is $2.5+0.5=3$ km/h. We need to find the total time, which is walking time + $t$. Walking time is $9/3=3$ hours. Total time is 3 hours + 24 minutes. Convert 3 hours to minutes: 180 minutes. Total time: $180+24=204$ minutes. So, the walk takes 204 minutes, including the coffee shop time. \textbf{Final Answer} The walk takes $\boxed{204}$ minutes.
\\

{\color[HTML]{2EA121}\faCogs\ \textbf{\textsc{AttnPO}-1.5B (ours)}}: \\
Okay, let's try to solve this problem step by step. So, Aya walks 9 kilometers every morning. When she walks at a speed of $s$ km/h, it takes her 4 hours, including $t$ minutes in the coffee shop. When she increases her speed to $s+2$ km/h, the total time becomes 2 hours and 24 minutes, still including $t$ minutes in the coffee shop. We need to find the time when she walks at $s+\frac{1}{2}$ km/h. First, let's parse the information. When walking at speed $s$ km/h: Distance $=9$ km, Time taken walking + time in coffee shop $=4$ hours. So, time walking is $\frac{9}{s}$ hours, and time in coffee shop is $t$ minutes. But $t$ is given in minutes, so we need to convert 4 hours to minutes to keep units consistent. 4 hours $=240$ minutes. So, time walking + $t=240$ minutes. Similarly, when walking at $s+2$ km/h: Time taken walking + time in coffee shop $=2$ hours $24$ minutes $=2\times60+24=144$ minutes. So, time walking is $\frac{9}{s+2}$ hours, which is $\frac{9}{s+2}\times60$ minutes, and time in coffee shop is $t$ minutes. So, let's set up the equations. First case: $\frac{9}{s}\times60+t=240$. Second case: $\frac{9}{s+2}\times60+t=144$. We can subtract the first equation from the second to eliminate $t$. So, $\left[\frac{9}{s+2}\times60+t\right]-\left[\frac{9}{s}\times60+t\right]=144-240$. Simplify: $\frac{9\times60}{s+2}-\frac{9\times60}{s}=-96$. Factor out $9\times60$: $9\times60\left(\frac{1}{s+2}-\frac{1}{s}\right)=-96$. Compute $\frac{1}{s+2}-\frac{1}{s}=\frac{s-(s+2)}{s(s+2)}=\frac{-2}{s(s+2)}$. So, $9\times60\times\frac{-2}{s(s+2)}=-96$. Compute $9\times60=540$. So, $\frac{540\times(-2)}{s(s+2)}=-96$. Multiply both sides by $-1$: $\frac{540\times2}{s(s+2)}=96$. So, $\frac{1080}{s(s+2)}=96$. Thus, $s(s+2)=\frac{1080}{96}=11.25$. So, $s^2+2s-11.25=0$. Solving quadratic equation: $s=\frac{-2\pm\sqrt{4+45}}{2}=\frac{-2\pm\sqrt{49}}{2}=\frac{-2\pm7}{2}$. Since speed can't be negative, $s=\frac{5}{2}=2.5$ km/h. So, $s=2.5$ km/h. Now, find $t$. From first case: $\frac{9}{2.5}\times60+t=240$. Compute $\frac{9}{2.5}=3.6$. $3.6\times60=216$. So, $216+t=240\Rightarrow t=24$ minutes. Now, when walking at $s+\frac{1}{2}=2.5+0.5=3$ km/h. Time walking: $\frac{9}{3}=3$ hours $=180$ minutes. Time in coffee shop: $t=24$ minutes. Total time: $180+24=204$ minutes. \textbf{Final Answer} $\boxed{204}$

\end{tcolorbox}
\caption{This is a problem from AIME 2024, where TLMRE introduces unnecessary variables and follows an incorrect modeling idea before correcting itself, while \methodname avoids such detours and goes straight to the correct formulation, resulting in more concise reasoning.}
\label{case2}
\end{figure*}

\begin{figure*}
\centering
\small
\begin{tcolorbox}[size=title,opacityfill=0.1]
\noindent
\faUser\ \textbf{Question (MMLU)}: A group of hikers climbed from Salt Flats (elevation $-55$ feet) to Talon Bluff (elevation $620$ feet). What is the difference in elevation between Talon Bluff and Salt Flats? A. $565$ feet ; B. $575$ feet ; C. $665$ feet ; D. $675$ feet. Output the correct option (one of ABCD) in $\boxed{}$, such as $\boxed{A}$. \\

\faCogs\ \textbf{TLMRE}:
Okay, so there's this problem about hikers climbing from Salt Flats to Talon Bluff, and I need to find the difference in elevation between these two places. Let me see... Salt Flats are at $-55$ feet, and Talon Bluff is at $620$ feet. Hmm, the question is asking for the difference in elevation, so I think that means how much higher Talon Bluff is compared to Salt Flats. Wait, elevation can be positive or negative, so the difference would be the absolute value of the higher elevation minus the lower one. But let me clarify: the elevation of Salt Flats is $-55$ feet, which is below sea level, and Talon Bluff is at $+620$ feet, which is much higher. So the difference should be the total vertical distance between them. To find the difference, I can subtract the lower elevation from the higher one. So, $620$ feet minus $(-55$ feet). Subtracting a negative is like adding, so that would be $620+55$. Let me calculate that: $620+55$ is $675$. So the difference in elevation is $675$ feet. \textcolor{red}{Let me double-check: starting at $-55$, you go up to $620$. The total change is $620-(-55)=620+55=675$. Yep, that seems right. Looking at the options, D is $675$ feet. So the correct answer should be D. \textbf{Final Answer} $\boxed{D}$ }

{\color[HTML]{2EA121}\faCogs\ \textbf{\textsc{AttnPO}-1.5B(ours)}}: \\
Okay, so I have this problem about hikers climbing from Salt Flats to Talon Bluff. The elevation of Salt Flats is $-55$ feet, and Talon Bluff is $620$ feet. I need to find the difference in elevation between the two places. Hmm, let's see. Elevation difference is usually the absolute value of the difference between the two points. So, I should subtract the lower elevation from the higher one. But one is negative, so I need to handle that carefully. Elevation of Salt Flats is $-55$ feet, which is below sea level. Talon Bluff is $620$ feet, which is much higher. So, the difference would be $620-(-55)$. Subtracting a negative is like adding, so $620+55$ equals $675$ feet. So the difference in elevation is $675$ feet. Looking at the options, D is $675$ feet. \textbf{Final Answer} $\boxed{D}$

\end{tcolorbox}
\caption{This is a simple calculation problem from MMLU, and compared to \methodname, TLMRE exhibits redundant self-reflection.}
\label{case3}
\end{figure*}

\begin{figure*}
\centering
\small
\begin{tcolorbox}[size=title,opacityfill=0.1]
\noindent
\faUser\ \textbf{Question (MMLU)}: Which of the jovian planets have rings? A. Neptune B. Uranus C. Saturn D. all of the above.  Output the correct option (one of ABCD) in $\boxed{}$, such as $\boxed{A}$. \\

\faCogs\ \textbf{TLMRE}:
Okay, so I need to figure out which jovian planets have rings. The options are Neptune, Uranus, Saturn, and all of the above. Let me start by recalling what I know about each of these planets. First, the jovian planets are the gas giants, right? So they're larger than the gas giants. The main ones are Jupiter, Saturn, Uranus, and Neptune. Now, I remember that gas giants do have rings, but I'm not entirely sure about each one. Let me think about each planet: Neptune: I believe Neptune has rings. It's a large planet, and like many gas giants, it has a ring system. \textcolor{red}{I think it's called the Great Ring System because it's a big, connected ring system around it.} Uranus: Uranus also has rings. It's another gas giant, and I recall that Uranus has a ring system that's quite prominent. There's a ring system that's pretty extensive around it. Saturn: Saturn is the largest of the gas giants. It has a ring system, but it's a bit more diffuse. \textcolor{red}{I think the rings are called Cassini's rings, which are not as tightly packed as Neptune's and Uranus's.} So all three: Neptune, Uranus, and Saturn have rings. Therefore, the correct answer should be D, all of the above.

{\color[HTML]{2EA121}\faCogs\ \textbf{\textsc{AttnPO}-1.5B(ours)}}: \\
Okay, so I have this question about jovian planets and their rings. The options are Neptune, Uranus, Saturn, and all of the above. Hmm, I remember that jovian planets are gas giants, so they're the gas giants themselves. Let me think about each option. Neptune: I recall that Neptune has a ring system. It's a gas giant with a lot of ice and rock around it. So Neptune has rings. Uranus: Similarly, Uranus is another gas giant. I think it also has a ring system. So Uranus also has rings. Saturn: Saturn is a gas giant too. It has rings as well. So Saturn has rings. Therefore, all three—Neptune, Uranus, Saturn—have rings. So the correct answer should be D, all of the above.

\end{tcolorbox}
\caption{This is a simple knowledge-based question from MMLU. The response from TLMRE introduces information that is not directly necessary for solving the problem, whereas \methodname does not.}
\label{case4}
\end{figure*}

\end{document}